\documentclass[11pt, letterpaper]{article}
\usepackage{graphicx}
\usepackage{amsfonts}
\usepackage[margin=1in]{geometry}
\usepackage{pgfplots}
\usepackage{pgfplotstable}
\usepackage{caption}
\usepackage{subcaption}
\pgfplotsset{compat=1.17}
 \usepackage{url}
\usepackage{quantikz}

\usepackage[numbers,sort&compress]{natbib}

\usepackage{tikz}
\usetikzlibrary{backgrounds,fit,decorations.pathreplacing,calc}
\usetikzlibrary{circuits.ee.IEC}
\usepackage{amsmath, bm, braket}

\usepackage{bm}  
\usepackage{amsmath}       
\usepackage{amssymb}       
\usepackage{bm}   

\usepackage{setspace}
\usepackage{parskip}
\usepackage{lmodern} 

\begin{document}

\title{Quantum Temporal Fusion Transformer}
\author{Krishnakanta Barik\footnote{krishnakanta\_r@isical.ac.in}, Goutam Paul\footnote{goutam.paul@isical.ac.in} \\  \textsuperscript{*}Cryptology and Security Research Unit, Indian Statistical Institute, Kolkata, India \\ \textsuperscript{†}Electronics and Communication Sciences Unit, Indian Statistical Institute, Kolkata, India}

\date{\today}

\maketitle


\begin{abstract}
The \textit{Temporal Fusion Transformer} (TFT), proposed by Lim \textit{et al.}, published in \textit{International Journal of Forecasting} (2021), is a state-of-the-art attention-based deep neural network architecture specifically designed for multi-horizon time series forecasting. It has demonstrated significant performance improvements over existing benchmarks. 
In this work, we introduce the Quantum Temporal Fusion Transformer (QTFT), a quantum-enhanced hybrid quantum-classical architecture that extends the capabilities of the classical TFT framework. 
The core idea of this work is inspired by the foundation studies, \textit{The Power of Quantum Neural Networks} by Amira Abbas \textit{et al.} and \textit{Quantum Vision Transformers} by El Amine Cherrat \textit{et al.}, published in \textit{ Nature Computational Science} (2021) and \textit{Quantum} (2024), respectively.
A key advantage of our approach lies in its foundation on a variational quantum algorithm, enabling implementation on current noisy intermediate-scale quantum (NISQ) devices without strict requirements on the number of qubits or circuit depth. 
Our results demonstrate that QTFT is successfully trained on the forecasting datasets and is capable of accurately predicting future values.   
In particular, our experimental results on two different datasets display that the model outperforms its classical counterpart in terms of both training and test loss. These results indicate the prospect of using quantum computing to boost deep learning architectures in complex machine learning tasks.

\end{abstract}

\newpage

\section{Introduction}

\begin{figure*}[ht]
\begin{center}
\resizebox{0.6\textwidth}{!}{
\begin{tikzpicture}
\tikzstyle{surround} = [fill=lime!10,thick,draw = black,rounded corners = 2mm]

\draw[->](-2, 0)--(13,0);
\draw[->](0, -1)--(0,6);
\draw[dashed](6, 0)--(6,5);
\draw node at (12.5,-.2) {Time};
\draw node[rotate=90] at (-.3,4) {Predicts Variables};
\draw node at (6,5.2) {Forecast Time (t)};
\filldraw (11,3) circle (1.5pt);
\filldraw (10,3.5) circle (1.5pt);
\filldraw (9,2.5) circle (1.5pt);
\filldraw (8, 1.75) circle (1.5pt);
\filldraw (7,1.5) circle (1.5pt);

\filldraw (5,1.35) circle (1.5pt);
\filldraw (4,1.25) circle (1.5pt);
\filldraw (3,.75) circle (1.5pt);
\filldraw (2,1) circle (1.5pt);
\filldraw (1,.5) circle (1.5pt);

\draw (11,3)--(10,3.5)--(9,2.5);
\draw[dashed] (9,2.5)--(8, 1.75);
\draw (8, 1.75)--(7,1.5);
\draw (3,.75)--(2,1)--(1,.5);
\draw[dashed] (4,1.25)--(3,.75);
\draw (5,1.35)--(4,1.25);
\draw (7,1.5)--(5,1.35);

\draw (1,.2)--(1,-.2);
\draw (2,.2)--(2,-.2);
\draw (3,.2)--(3,-.2);
\draw (4,.2)--(4,-.2);
\draw (5,.2)--(5,-.2);
\draw node at (1,-.4) {$t-k$};
\draw node at (5,-.4) {$t$};

\draw (7,.2)--(7,-.2);
\draw (8,.2)--(8,-.2);
\draw (9,.2)--(9,-.2);
\draw (10,.2)--(10,-.2);
\draw (11,.2)--(11,-.2);
\draw node at (7,-.4) {$t+1$};
\draw node at (11,-.4) {$t+\tau_{\max}$};

\draw (1,.5)-- (3, 3);
\draw (2,1)-- (3, 3);
\draw (3,.75)-- (3, 3);
\draw (4,1.25)-- (3, 3);
\draw (5,1.35)-- (3, 3);
\draw node at (3,3.2) {Past Targets};

\draw (11,3)-- (9.5,4.3);
\draw (10,3.5)-- (9.5,4.3);
\draw (9,2.5)-- (9.5,4.3);
\draw (8, 1.75)-- (9.5,4.3);
\draw (7,1.5)-- (9.5,4.3);
\draw node at (9.5,4.5) {Future Predictions};

\draw[->](1.6, -.7)--(1.6,-.5);
\draw[->](4.4, -.7)--(4.4,-.5);
\draw (1.3, -.7) rectangle (4.7,-1.3);
\draw node at (3,-1) {Know Past Inputs};

\draw[->](7.6, -.7)--(7.6,-.5);
\draw[->](10.2, -.7)--(10.2,-.5);
\draw (7.2, -.7) rectangle (10.8,-1.3);
\draw node at (9,-1) {Know Future Inputs};


\end{tikzpicture}}
\caption{Illustration of multi-horizon forecasting.
The X-axis represents the time steps (sliding window), while the Y-axis represents the target variables to be predicted. The forecast time point is denoted as $t$. The model uses historical data from $t-k$ to $t$ to predict the selected variable over the future horizon, from $t$ to $t + \tau_{\max}$.}
\label{fig:multi-horizon forecasting}
\end{center}
\end{figure*}
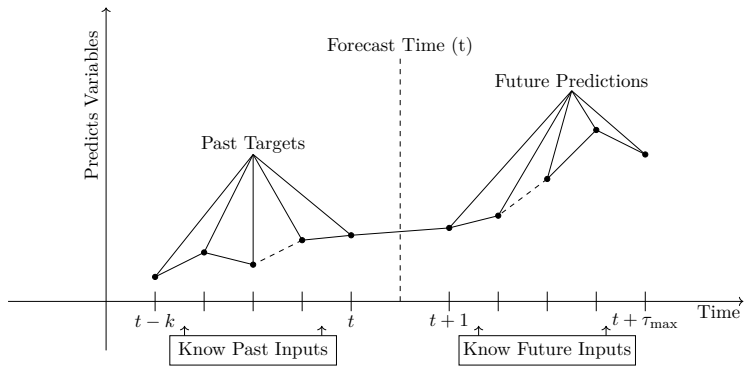

Multi-horizon forecasting is a time series forecasting framework~\cite {box2015time}, in which a model predicts interesting variables over multiple future time steps. Unlike standard time series forecasting, which predicts variables one step ahead, multi-horizon forecasting predicts variables for several future time points, thereby capturing predictions across the entered future path.  
Multi-horizon forecasting has broad applications in the real world, including healthcare~\cite{lim2018forecasting, zhang2018multi,piccialli2021artificial}, financial \cite{kroujiline2016forecasting, capistran2010multi}, retail \cite{bose2017probabilistic, courty1999timing}. Figure~\ref{fig:multi-horizon forecasting} provides an overview of the overall architecture of multi-horizon forecasting.
Multi-horizon forecasting is based on multiple data sources such as time-independent fixed features (e.g., the store's location), known future information (e.g., an upcoming holiday), and comprehensive historical data (e.g., customer price trade).
Without understanding the relationships among these diverse data sources, multi-horizon time series forecasting is a challenging task.        

There exist various architectures based on Recurrent Neural Networks (RNNs)~\cite{salinas2020deepar, rangapuram2018deep, wen2017multi} that have been designed for doing multi-horizon time series forecasting. Deep Neural Networks (DNNs) have been used extensively, and have performed compared to conventional time series models~\cite{alaa2019attentive, rangapuram2018deep, makridakis2020m4}. Recently, transformer-based models~\cite{li2019enhancing} have been proposed for time series forecasting and have demonstrated better performance. Yet these models occasionally underperform or fail when they handle various kinds of inputs that frequently occur in multi-horizon forecasting~\cite{salinas2020deepar, wen2017multi, rangapuram2018deep, li2019enhancing}.
In the paper~\cite{lim2021temporal}, Lim et al. introduced a new model, the Temporal Fusion Transformer (TFT), a novel model for multi-horizon forecasting.  Building upon Deep Neural Networks (DNNs) and attention mechanisms~\cite{vaswani2017attention}, the TFT model demonstrates superior performance compared to existing models.

Quantum computing is a rapidly growing field in computer science that uses the harnessing of quantum bits (qubits) and the principles of quantum mechanics, such as entanglement and superposition, unlocking computational power beyond that of classical algorithms.
Several leading technology companies, including Google~\cite{arute2019quantum}, IBM~\cite{cross2018ibm}, and D-Wave~\cite{lanting2014entanglement}, have developed quantum computers that are accessible to the general public through cloud-based services. These advancements represent significant progress in making quantum computing more practical, pushing research and innovation across various scientific and industrial domains.
Quantum computing are capable for solving a classes of problems exponentially faster than existing classical computing~\cite{Harrow_2009, J__2022}. However, this magnificent speed-up vastly depends on the standard of the underlying quantum computer hardware. Quantum circuits with a large number of qubits or deep circuit depths are not reliably executed on current quantum devices, called Noisy Intermediate-Scale Quantum (NISQ) devices~\cite{preskill2018quantum} due to the presence of quantum error and noise~\cite{gottesman1997stabilizer, gottesman1998theory}. Therefore, it is significant to design quantum frameworks for execution on NISQ devices, ensuring better outcomes despite hardware limitations.

Quantum Machine Learning (QML)~\cite{biamonte2017quantum} is a combined domain of quantum computing and machine learning, which uses the strengths of quantum systems to enhance traditional machine learning tasks.
Quantum Variational Algorithms (VQAs)~\cite{cerezo2021variational, mitarai2018quantum, wecker2015progress, higgott2019variational} are one of the breakthrough innovations of quantum machine learning, providing a promising algorithm with the potential for applicability to NISQ devices.
A VQA is basically an appropriate quantum parametrized circuit, where gate parameters are adjustable and updated iteratively through a classical optimization process to find the solution to a given problem.
Since VQAs are an iterative optimization process, the quantum noise inherent in quantum devices can usually be effectively mitigated through the tunable parameters of the quantum circuit.
Consequently, VQAs are particularly suitable for implementation on the currently available NISQ devices. 
 
In this paper, we address the challenges of learning sequential data using Quantum Machine Learning techniques (QML)~\cite{biamonte2017quantum, dunjko2018machine, schuld2015introduction, huang2021power}. 
We propose a novel methodology for the practically feasible implementation of attention-based deep neural networks using variational quantum algorithms.
Specifically, we propose a Quantum version of Temporal Fusion Transformer (TFT), which is an attention-based deep neural network capable of learning from time series data and performing multi-horizon forecasting, using variational quantum algorithms.
The building block of our quantum-classical hybrid model is VQA, so the proposed model is efficiently implementable on current noise quantum hardware (NISQ devices).   
In the numerical simulation part, we implement a simplified quantum version of the TFT model, due to the limitations of existing quantum hardware. Our proposed quantum model outperforms its classical counterpart in terms of both training and test loss. 
However, in the future, as quantum computers overcome constraints, they also have the potential to deliver significantly better results for large-scale models.  
To the best of our knowledge, this work presents the first successful mapping of a large-scale classical learning model into a quantum learning framework with some potential advantage.

\textbf{Our contributions are summarized as follows}
\begin{enumerate}
    \item   We introduce, for the first time, quantum-enhanced 
            Gated Residual Network~\cite{lim2021temporal} and  Interpretable Multi-head Attention~\cite{lim2021temporal}.
    \item   We are the first to train and evaluate a
            quantum-enhanced Temporal Fusion Transformer (QTFT) model to perform multi-horizon time series forecasting.
    \item   We employ two distinct types of datasets for         experimental evaluation and demonstrate 
            improved results, indicating that our model has greater generalizability.  
\end{enumerate}

The remainder of this paper is as follows. 
First, in Section 2, we provide a brief review of the classical temporal fusion transformer, including an explanation of each component and details of the model architecture.
In Section 3, we introduce the variational quantum algorithm, the building block of our model.  
We discuss our main proposal in Section 4. This section explains the tools used in QTFT,  outlines model architecture, and describes the optimization procedure.
In Section 5, we present the implementation of our model, followed by the conclusion in Section 6.   

\section{Classical Temporal Fusion Transformer\label{CLASSICAL TEMPORAL FUSION TRANSFORMER}}
Throughout this discussion, for the sake of explanation and understanding, we consider the dataset of stores in retail and patients in healthcare. 
We use the same notation as the paper~\cite{lim2021temporal}.
There are three main input components of the Temporal Fusion Transformer (TFT): a set of static covariates $\bm{s} \in \mathbb{R}^{m_s}$, where $m_s$ be the dimension of static variables, time-dependent inputs $\bm{\chi}_t \in \mathbb{R}^{m_{\chi}}$, and corresponding scalar targets output $y_t$ at each time step $t$ between $0$ to $T$. 
Static covariates provide information that doesn’t change over time (e.g., store’s size). The time-dependent inputs separate into two categories: observed inputs $\bm{z}_t \in \mathbb{R}^{m_z}$, which can only measure them after they happen (e.g., weather), and know inputs $\bm{x}_t \in  \mathbb{R}^{m_x}$, that are known beforehand (e.g., holiday, voting day).

There is another term related to forecasting, called quantile forecasting, which is a technique that predicts an interval of the possible outputs rather than a single point output.
Let $f_{(.)}$ is the quantile-specific prediction model for the forecast horizon spans $\tau \in \{1, 2, \dots, \tau_{\max} \}$, and $k$ defines the size of the past information window.
Then $\hat{y}(q, t, \tau)$ is the predicted $q$-th sample quantile for the forecast $\tau $ time steps ahead at a time $t$, define as
\begin{equation*}
    \hat{y}(q, t, \tau) = f_q \left( \tau, y_{t-k:t},\bm z_{t-k:t},\bm x_{t-k:t+\tau},\bm s \right),
\end{equation*}
where, $ y_{t-k:t} = \{y_{t-k}, y_{t-k+1}, \dots y_t\}$ and similarly for $\bm z, \bm x$.

\subsection{Components}
The TFT model uses several components to learn the time series data for successful forecasting.

\subsubsection{Gated Residual Networks~\cite{lim2021temporal}}

The relationship between multi-dimensional inputs and target outputs is typically unknown in advance, making it challenging to estimate which input features are most relevant for prediction. 
It is challenging to handle this data in the context of non-linear and linear processing in the models to more accurately predict target values.  
The Gated Residual Networks (GRN) address this issue by combining a non-linear activation function and a residual connection for flexibility, applying non-linear functions where needed, and better handling of the data in the model.
Gated Residual Networks (GRN) play a crucial role in the Temporal Fusion Transformer (TFT) model. 

GRN receives two inputs: primary input $\bm{a}$ and optional input $\bm{c}$.
Fast, the primary input $\bm{a}$ and the optional input $\bm{c}$ are passed through a neural network with an Exponential Linear Unit (ELU)~\cite{clevert2015fast} activation function
\begin{equation*}
    \bm\eta_1 = \text{ELU}\left( \bm{W}_{1} \bm{a} + \bm{W}_{2}\bm{c} + \bm{b}_{12} \right),   
\end{equation*}
where $\bm{W}_{(.)}$ and $\bm{b}_{(.)}$ are denoted as the learnable weight matrix and bias vector, respectively.
The ELU would behave like an identity function when the input is positive, and for negative input, the  ELU would generate a constant output.
Next, the output $\bm\eta_1$ from the previous layer passes through another neural network without any activation function
\begin{equation*}
    \bm\eta_2 = \bm{W}_{3} \bm{\eta}_1 + \bm{b}_3.   
\end{equation*}

\begin{figure*}[ht]
\begin{center}
\resizebox{0.6\textwidth}{!}{
\begin{tikzpicture}
\tikzstyle{surround} = [fill=lime!10,thick,draw = black,rounded corners = 2mm]

\draw node at (.4,2.2) {$\bm{a}$};
\draw node at (.4,0.2) {$\bm{c}$};
\draw   [->] (0,0)--(1,0);
\draw   [->] (0, 2)--(1,2);
\draw (1, 2.5) rectangle (3,-.5);
\filldraw (1.5,1.5) circle (2pt);
\filldraw (1.5,.5) circle (2pt);

\filldraw (2.5,2) circle (2pt);
\filldraw (2.5,1) circle (2pt);
\filldraw (2.5,0) circle (2pt);

\draw (1.5,1.5)--(2.5,2);
\draw (1.5,1.5)--(2.5,1);
\draw (1.5,1.5)--(2.5,0);

\draw (1.5,.5)--(2.5,2);
\draw (1.5,.5)--(2.5,1);
\draw (1.5,.5)--(2.5,0);

\draw node at (2,-0.3){$\mathbf{W}_{1}, \mathbf{W}_{2} $};

\draw[->] (3,1)--(4,1);
\draw (4, 2.5) rectangle (5.5,-.5);

\filldraw (4.25,1.5) circle (2pt);
\filldraw (4.25,.5) circle (2pt);

\filldraw (5.25,2) circle (2pt);
\filldraw (5.25,1) circle (2pt);
\filldraw (5.25,0) circle (2pt);

\draw (4.25,1.5)--(5.25,2);
\draw (4.25,1.5)--(5.25,1);
\draw (4.25,1.5)--(5.25,0);

\draw (4.25,.5)--(5.25,2);
\draw (4.25,.5)--(5.25,1);
\draw (4.25,.5)--(5.25,0);

\draw node at (4.75,-0.3){$\mathbf{W}_{3}$};

\draw[->] (5.5,1)--(7,1);
\draw (7, 2.5) rectangle (9,-.5);

\filldraw (7.5,2) circle (2pt);
\filldraw (7.5,1.5) circle (2pt);

\filldraw (8.5,2.2) circle (2pt);
\filldraw (8.5,1.3) circle (2pt);

\draw (7.5,2)--(8.5,2.2);
\draw (7.5,2)--(8.5,1.3);

\draw (7.5,1.5)--(8.5,2.2);
\draw (7.5,1.5)--(8.5,1.3);

\filldraw (7.5,.7) circle (2pt);
\filldraw (7.5,.2) circle (2pt);

\filldraw (8.5,.9) circle (2pt);
\filldraw (8.5,-.0) circle (2pt);

\draw (7.5,.7)--(8.5,.9);
\draw (7.5,.7)--(8.5,-.0);

\draw (7.5,0.2)--(8.5,.9);
\draw (7.5,0.2)--(8.5,-.0);

\draw node at (8,-0.3){$\mathbf{W}_{4}, \mathbf{W}_{5}$};
\draw   [->] (9,1)--(10,1);
\draw (10, 2.5) rectangle (11.5,-.5);

\draw (10.6,1)--(10.9,1);
\draw (10.75,.85)--(10.75,1.15);


\draw   [->] (11.5,1)--(12.5,1);
\draw   (.75, 2)--(.75,3);
\draw   (.75, 3)--(10.75,3);
\draw   [->](10.75, 3)--(10.75,2.5);
\end{tikzpicture}}
\caption{Generic architecture for Gated Residual Networks (GRNs). The input $\bm a$ represents the primary input, and $\bm c$ is an optional external context vector. $\mathbf{W}_{1}, \mathbf{W}_{2}$ is a dense layer (neural network) followed by an ELU activation function. $\mathbf{W}_{3}$ is another dense layer without activation function. $\mathbf{W}_{4}, \mathbf{W}_{5}$ represented the Gated Linear Unit (GLU) operation. Final block performance residual connection (add) and layer normalization.}
\label{fig: GRN}
\end{center}
\end{figure*}
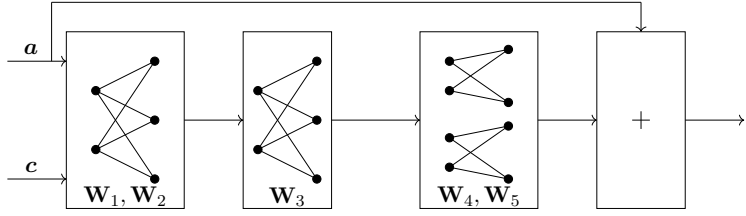
 
Now $\bm\eta_2$ are fitted into gating layers based on Gated Linear Units (GLUs)~\cite{dauphin2017language} to selectively deactivate parts of the model that are unnecessary for a specific dataset. The GLU is defined as follows
\begin{equation*}
   \bm\eta_3 = \text{GLU}(\bm{\eta}_2) = \sigma\left( \bm{W}_{4} \bm{\eta}_2 + \bm{b}_{4} \right) \odot \left( \bm{W}_{5} \bm{\eta}_2 + \bm{b}_{5} \right),
\end{equation*}
where $\sigma(.)$ denote as sigmoid activation function and $\odot$ is the element-wise Hadamard product.
Finally, the input $\bm{a}$ is combined with $\bm{\eta}_3$ through a residual connection, and the result is refined through a layer normalization step~\cite{ba2016layer} as below, ensuring stable and consistent activations,
\begin{equation*}
    \mathrm{GRN}(\bm{a}, \bm{c}) = \text{LayerNorm} \bigl(\bm{a} + \bm\eta_3 \bigr).
\end{equation*}

\subsubsection{Variable Selection Networks~\cite{lim2021temporal}}

Variables play a crucial role in multi-horizon forecasting. While certain variables provide significant predictive power, others may introduce unnecessary noise without impacting performance. Therefore, identifying and distinguishing the most appropriate variables is a challenging task for improving overall model effectiveness.
To address the issues, variable selection networks, a learnable model, provide an effective solution for efficiently handling multiple variables in the dataset.
For better mathematical representation, categorical variables are encoded using entity embedding~\cite{gal2016theoretically} of dimension $d_{model}$,  while continuous variables are transformed linearly with the same dimension $d_{model}$.  
Variational selection networks are applied separately to all three types of inputs - static, past, and future. Here, discuss variational selection networks for past inputs; the same structure is applied to both static and future inputs.

Let the encoded past input of $j$-th variable at time $t$ be denoted by $\bm\xi_t^{(j)} \in \mathbb{R}^{d_{model}}$. Then encoded past inputs data at time $t$ are concatenated, and denoted as a flattened vector 
\begin{equation*}
    \bm\Xi_t = \left[ \bm\xi_{t}^{(1)^T}, \ldots, \bm\xi_t^{(m_\chi)^T} \right]^T.
\end{equation*}
After, both $\bm\Xi_t$ and an external context vector $\bm c_s$, obtained from a static covariate encoder (discuss later), are passed through GRN, followed by a softmax layer~\cite{wang2018high}
\begin{equation*}
    \bm{v}_{\chi_t} = \text{Softmax} \left( \text{GRN} \left( \bm{\Xi}_t, \bm{c}_s \right) \right),
\end{equation*}
where the softmax function is defined as
\begin{align*}
    &\text{Softmax}\left(w_1, w_2, \dots, w_k\right) = \\
    &\left (\frac{e^{w_1}}{\sum_{i = 1}^k e^{w_{i}}}, \frac{e^{w_2}}{\sum_{i = 1}^k e^{w_{i}}}, \dots, \frac{e^{w_k}}{\sum_{i = 1}^k e^{w_{i}}}\right),
\end{align*}
for any $(w_1, w_2, \dots, w_k) \in \mathbb{R}^k$. $\bm{v}_{\chi_t}$ is an $m_{\chi}$ dimensional vector, called variable selection weights. 
At each time step $t$, another GRN layer is applied to encoded input $\bm\xi_{t}^{(j)}$, for all $j \in [0, m_{\chi}]$
\begin{equation*}
    \tilde{\bm{\xi}}_t^{(j)} = \text{GRN}\left( \bm{\xi}_t^{(j)} \right),
\end{equation*}
where $\tilde{\bm{\xi}}_t^{(j)}$ is called  processed feature vector. 
The final output of the variable selection network is a weighted sum of processed feature vectors, where the weights are given by the variable selection weights
\begin{equation*}
    \tilde{\bm{\xi}}_t = \sum_{j=1}^{m_{\chi}} v_{\chi_t}^{(j)}\tilde{\bm{\xi}}_t^{(j)},
\end{equation*}
where $v_{\chi t}^{(j)}$ is $j$-th component of the vector $\bm v_{\chi_t}$.

\subsubsection{Static Covariate Encoders~\cite{lim2021temporal}}

Static variables play a crucial role in time series forecasting, as different components of models utilize them in various forms. 
Specifically, there are three main places of the TFT model where four distinct context vectors are required to improve predictive accuracy. The context vectors $\bm c_s, \bm c_e,\bm c_c,\bm c_h$ are generated by a static covariate encoder using separate GRN encoders (different by parameters). Each encoder takes the fixed input $\tilde{\bm{\xi}}$, which is the output of the static variable selection network 
\begin{equation*}
    \bm{c}_{j} = \text{GRN}(\tilde{\bm{\xi}}), \quad j \in \{ s,e, c, h \}.
\end{equation*} 

\subsubsection{Interpretable Multi-Head Attention~\cite{lim2021temporal}}

The attention mechanism~\cite{vaswani2017attention, li2019enhancing} is an important tool for capturing long-term relationships between different elements in the input data.
We provide a general framework for applying the attention mechanism across different domains; in the context of the TFT,  we specifically incorporate it within the temporal self-attention layer.

Let $\bm S \in \mathbb{R}^{N \times d}$ be the matrix representing the input vectors. Let $\bm W_q, \bm W_k \in \mathbb{R}^{d \times d_{attn}}$  and $\bm W_v \in \mathbf{R}^{d \times d_{attn}}$ be learnable parameter matrices used to project the input into query $\bm Q$, key $\bm K$, and value $\bm V$ spaces, respectively, i.e., $\bm Q = \bm S\bm W_Q $, $\bm K = \bm S\bm W_K $, $\bm V = \bm S\bm W_v $.
The output of the attention operation is defined as
\begin{equation*}
    \text{Attention}(\bm Q,\bm K,\bm V) = A(\bm Q, \bm K) \bm V,
\end{equation*}
where $A(\bm Q, \bm K) = \text{softmax}\left(\frac{\bm Q \bm K^T}{\sqrt{d_{attn}}}\right)$.
Multi-head attention, introduced in~\cite{vaswani2017attention}, improves the learning capacity of the model by enabling it to jointly attend (different heads) to information from different representation subspaces at various positions of the given input data. 
If the number of attention heads is $ m_H$, then the output of the multi-head attention mechanism is given by 
\begin{equation*}
    \text{Multi-head}(\bm Q, \bm K, \bm V) = \left[\text{Attention}\left (\bm Q^{(1)}, \bm K^{(1)}, \bm V^{(1)}\right), \dots,  \text{Attention} \left (\bm Q^{(m_H)},\bm  K^{(m_H)}, \bm V^{(m_H)}\right )\right ] \bm W_H, 
\end{equation*}
where $\bm Q^{(h)}, \bm K^{(h)}, \bm V^{(h)}$ are weights for queries, key, and value projections for the $h$-th attention head, and $ \bm W_H$ is the matrix used to combine the concatenated outputs of all attention heads.

In a multi-head attention mechanism, the value vectors ($\bm V^{(.)}$)  play a crucial role in determining the importance of specific features. Different value vectors are used in different heads; they may fail to prioritize certain features consistently.
In contrast, sharing one fixed value vector in all heads and additive aggregation of all heads increase the model’s capacity efficiently. This approach is known as Interpretable Multi-head Attention~\cite{lim2021temporal} 
\begin{equation*}
\text{InterpretableMultiHead}(\bm{Q}, \bm{K}, \bm{V}) = \tilde{\bm{H}}\bm W_{\tilde{H}},   
\end{equation*}
where
\begin{equation*}
    \tilde{\bm{H}}= \frac{1}{m_H} \sum_{h=1}^{m_H} \text{Attention}\left( \bm{Q}^{(h)}, \bm{K}^{(h)}, \bm{V}\right ),
\end{equation*}
and $\mathbf W_{\tilde{H}}$ is applied as a final linear projection. 

\subsection{Temporal Fusion Transformer}
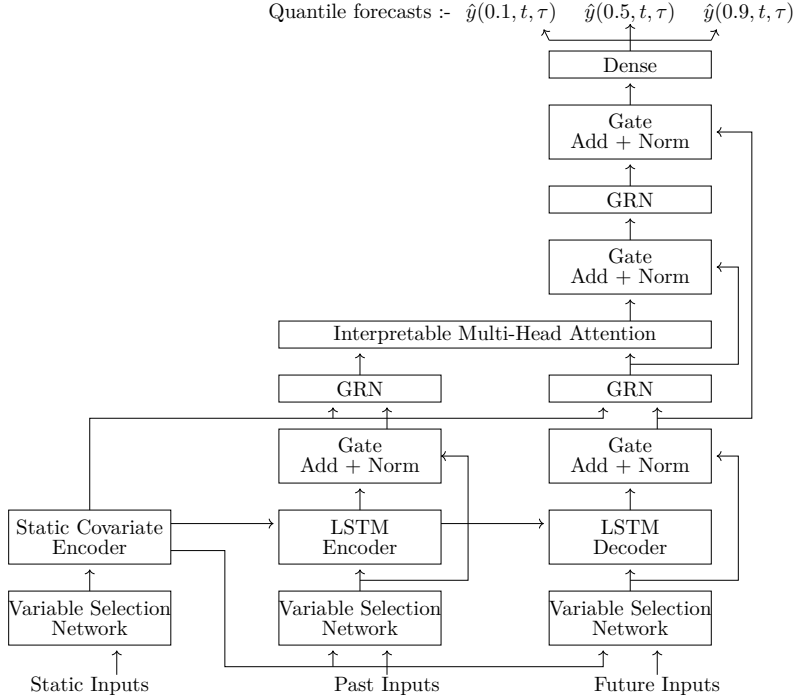
\begin{figure*}[ht]
\begin{center}
\resizebox{.65\textwidth}{!}{
\begin{tikzpicture}
\tikzstyle{surround} = [fill=lime!10,thick,draw = black,rounded corners = 2mm]

\draw (0, 1) rectangle (3, 0);
\draw node at (1.5, 0.5) {  \shortstack{Variable Selection\\Network}};
\draw node at (1.5, -.75) {  Static Inputs};
\draw[->] (2,-.55)--(2,-.1);

\draw (5, 1) rectangle (8, 0);
\draw node at (6.5, 0.5) {  \shortstack{Variable Selection\\Network}};
\draw node at (7, -.75) {  Past Inputs};
\draw[->] (7,-.55)--(7,-.1);

\draw (10, 1) rectangle (13, 0);
\draw node at (11.5, 0.5) {  \shortstack{Variable Selection\\Network}};

\draw node at (12, -.75) {  Future Inputs};
\draw[->] (12,-.55)--(12,-.1);

\draw (0, 2.5) rectangle (3, 1.5);
\draw node at (1.5, 2) {  \shortstack {Static Covariate \\Encoder}};
\draw[->] (1.5,1)--(1.5,1.4);

\draw (5, 2.5) rectangle (8, 1.5);
\draw node at (6.5, 2) {  \shortstack {LSTM \\Encoder}};
\draw[->] (6.5,1)--(6.5,1.4);

\draw (10, 2.5) rectangle (13, 1.5);
\draw node at (11.5, 2) {  \shortstack {LSTM\\ Decoder}};
\draw[->] (11.5,1)--(11.5,1.4);

\draw[->] (3, 1.75)--(4,1.75)--(4, -0.4)--(11, -0.4)--(11, -0.1);
\draw[->] (6, -0.4)--(6, -0.1);

\draw[->] (3, 2.25)--(4.9, 2.25);
\draw[->] (8, 2.25)--(9.9, 2.25);

\draw (5, 4) rectangle (8,3);
\draw node at (6.5, 3.5) {  \shortstack {Gate\\ Add + Norm}};
\draw (10, 4) rectangle (13,3);
\draw node at (11.5, 3.5) {  \shortstack {Gate\\ Add + Norm}};
\draw[->] (6.5, 2.5)--(6.5, 2.9);
\draw[->] (11.5, 2.5)--(11.5, 2.9);
\draw[->] (6.5, 1.2)--(8.5, 1.2)--(8.5, 3.5)--(8, 3.5);
\draw[->] (11.5, 1.2)--(13.5, 1.2)--(13.5, 3.5)--(13.1, 3.5);
\draw (5, 4.5) rectangle (8,5);
\draw node at (6.5, 4.75) {  {GRN} };
\draw (10, 4.5) rectangle (13,5);
\draw node at (11.5, 4.75){  {GRN} };
\draw[->] (7, 4)--(7, 4.4);
\draw[->] (12, 4)--(12, 4.4);

\draw[->] (1.5, 2.5)--(1.5, 4.2)--(11, 4.2)--(11, 4.4);
\draw[->] (6, 4.2)--(6, 4.4);
\draw (5, 5.5) rectangle (13,6);
\draw node at (9, 5.75) {  {Interpretable Multi-Head Attention} };
\draw[->] (6.5, 5)--(6.5, 5.4);
\draw[->] (11.5, 5)--(11.5, 5.4);

\draw (10, 6.5) rectangle (13,7.5);
\draw node at (11.5, 7) {  \shortstack {Gate\\ Add + Norm}};
\draw[->] (12, 4.2)--(13.75, 4.2)--(13.75, 9.5)--(13.1, 9.5);
\draw (10, 8) rectangle (13,8.5);
\draw node at (11.5, 8.25){  {GRN} };
\draw (10, 9) rectangle (13,10);
\draw node at (11.5, 9.5) {  \shortstack {Gate\\ Add + Norm}};
\draw[->] (11.5, 5.2)--(13.5, 5.2)--(13.5, 7)--(13.1, 7);
\draw[->] (11.5, 6)--(11.5, 6.4);
\draw[->] (11.5, 7.5)--(11.5, 7.9);
\draw[->] (11.5, 8.5)--(11.5, 8.9);
\draw (10, 10.5) rectangle (13,11);
\draw node at (11.5, 10.75){  {Dense}};
\draw[->] (11.5, 10)--(11.5, 10.4);
\draw[->] (11.5, 11)--(11.5, 11.5);
\draw (10, 11.2)--(13, 11.2 );
\draw[->] (10, 11.2)--(9.9, 11.4);
\draw[->] (13, 11.2)--(13.1, 11.4);

\draw node at (6.5, 11.7){  {Quantile forecasts :-}};

\draw node at (9.3, 11.7){  {$\hat{y}(0.1, t, \tau)$}};
\draw node at (11.5, 11.7){  {$\hat{y}(0.5, t, \tau)$}};
\draw node at (13.7, 11.7){  {$\hat{y}(0.9, t, \tau)$}};
\end{tikzpicture}} 
\caption{ TFT architecture. TFT processes three types of inputs: static inputs, time-dependent past inputs, and prior known future inputs. The gated residual network facilitates the flexibility of information either through skip connections or via gated linear unit layers. The variable selection network dynamically identifies the most valuable features from the input data. LSTM layers capture local sequential dependencies, while interpretable multi-head attention enables the combining of information across all time steps.}
\label{fig: TFT}
\end{center}
\end{figure*}

Figure~\ref{fig: TFT} shows a high-level architecture of TFT, with individual layers explained in detail in the subsequent section.

\subsubsection{Locality Enhancement with Sequence-to-Sequence Layer}

In time series data, points such as anomalies, change-points, or cycles are detected by comparing values against their local context. Incorporating features that extract local patterns, instead of just individual points, can enhance the performance of attention-based models. For instance, one method applies a single convolution layer to extract local patterns. This approach, however, would not perform well if there are variable amounts of past and future input data.
The following describes the process of locality enhancement for input time series data using a sequence-to-sequence layer to handle these differences.

For outputs $\bm{\tilde\xi}_{t-k:t}$ from variable selection network, corresponding to past inputs, are passed through an LSTM~\cite{hochreiter1997long} encoder, while the outputs $\bm {\tilde\xi}_{t+1:t+\tau_{\max}}$ from variable selection network, corresponding feature inputs, are passed through LSTM decoder.
The cell state and hidden state of the first LSTM in the layer are initialized using the context vectors $\bm c_c$ and $\bm c_e$, respectively, which are obtained from static covariate encoders.
The outputs form this layer are denoted as $\bm \phi(t, -k), \ldots, \bm \phi(t, \tau_{\max})$.
The final outputs of this layer are  derived using Gated Linear Units (GLUs), applied through a residual connection, followed by layer normalization  
\begin{equation*} 
    \tilde{\bm \phi}(t, n) = \text{LayerNorm}\left(
    \bm {\tilde{\xi}}_{t+n} + \text{GLU}(\bm \phi(t, n))
    \right),  
\end{equation*}
where $ n \in [-k, \tau_{\max}]$.

\subsubsection{Static Enrichment Layer}

Temporal dynamics are significantly influenced by static metadata, and the static enrichment layer enhances these temporal features.
Specifically, the static enrichment layer applies a GRN to the output locality enhancement $\bm {\tilde{\phi}}(t, n)$, along with context vector $\bm c_e$ from the static covariate encoder.

\begin{equation*}
    \bm \theta(t, n) = \text{GRN}\bigl(
    \tilde{\bm \phi}(t, n),\, \bm c_e
\bigr), 
\end{equation*}
where  $n \in [-k, \tau_{\max}]$.

\subsubsection{Temporal Self-Attention Layer}
The long-range dependencies in the  TFT model are efficiently captured by the self-attention layer.  
The layer operates as follows.
Let $\bm \Theta(t) = [\bm \theta(t, -k), \ldots, \bm \theta(t, \tau_{max})]^T$ denote the matrix formed by stacking the outputs of the static enrichment layer. 
Subsequently, an interpretable multi-head attention mechanism is applied to $\bm \Theta(t)$
\begin{equation*}
    \bm B(t) = \text{InterpretableMultiHead}(\bm \Theta(t), \bm \Theta(t), \bm \Theta(t)),
\end{equation*}
where $\bm B(t) = [\bm \beta(t, -k), \ldots, \bm \beta(t, \tau_{\max})]$ represents the output of the interpretable multi-head attention mechanism.
A gating layer (GLU) is also included as the final component of this Layer to improve training efficiency
\begin{equation*}
    \bm \delta(t, n) = \text{LayerNorm}\big(\bm \theta(t, n) + \text{GLU}(\bm \beta(t, n))\big), 
\end{equation*}
where $n \in [-k, \tau_{\max}]$.
\subsubsection{Position-Wise Feed-Forward Layer}
In this layer, a non-linear module GRN is applied to the outputs of the temporal self-attention layer
\begin{equation*}
    \bm\psi(t, n) = \text{GRN} \big(\bm \delta(t, n) \big).
\end{equation*}
Additionally, a gated (GLU) residual connection is included via a direct pathway to the sequence-to-sequence layer
\begin{equation*}
    \tilde{\bm\psi}(t, n) = \text{LayerNorm}\left( \tilde{\bm\phi}(t, n) + \text{GLU}(\bm\psi(t, n)) \right), 
\end{equation*}
where $n \in [-k, \tau_{\max}]$.
\subsubsection{Quantile Outputs}
In many real-world cases, instead of predicting a single point estimate,  providing prediction intervals is valuable for optimizing decision-making and managing risk,  as it captures the likely best- and worst-case outcomes that the target variable can take. Quantile forecasting does this job by applying linear transformations to the output of the position-wise feed-forward layer
\begin{equation*}
    \hat{y}(q, t, \tau) = \bm{W}_q \, \tilde{\bm{\psi}}(t, \tau) + b_q,
\end{equation*}
where $\bm W_q$,  $b_q$
are the learnable coefficients corresponding to the specified quantile $q$, and since forecasts are only of interest for future time steps, $\tau \in [1, \tau_{\max}]$.

\section{Variational Quantum Algorithm\label{VARIATIONAL QUANTUM ALGORITHM}}
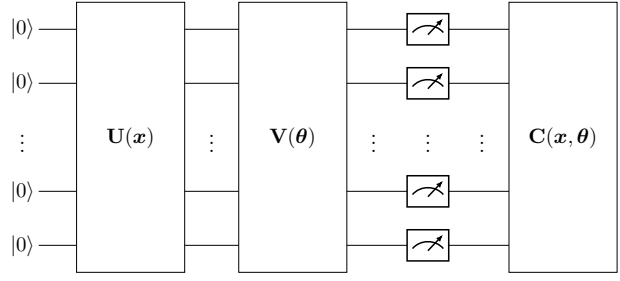
\begin{figure*}[ht]
\begin{center}
\resizebox{0.5\textwidth}{!}{
\begin{tikzpicture}
\tikzstyle{surround} = [fill=lime!10,thick,draw = black,rounded corners = 2mm]
\draw node at (0,0) {$\ket{0}$};
\draw node at (0,1) {$\ket{0}$};
 \draw node at (0,2) {$\vdots$};
\draw node at (0,3) {$\ket{0}$};
\draw node at (0,4) {$\ket{0}$};
\draw (0.3,0)--(1,0);
\draw (0.3,1)--(1,1);
\draw (0.3,3)--(1,3);
\draw (0.3,4)--(1,4);

\draw (1, 4.5) rectangle (3,-.5);
\draw node at (2, 2) {$\mathbf{U}(\bm x)$};

\draw (3,0)--(4,0);
\draw (3,1)--(4,1);
\draw (3,3)--(4,3);
\draw (3,4)--(4,4);

\draw node at (3.5,2) {$\vdots$};

\draw (4, 4.5) rectangle (6,-.5);
\draw node at (5, 2) {$\mathbf{V}(\bm \theta)$};

\draw (6,0)--(9,0);
\draw (6,1)--(9,1);
\draw (6,3)--(9,3);
\draw (6,4)--(9,4);

\draw node at (6.5,2) {$\vdots$};

\node[meter] (meter) at (7.5,0) {};
\node[meter] (meter) at (7.5,1) {};
\node[meter] (meter) at (7.5,3) {};
\node[meter] (meter) at (7.5,4) {};

\draw node at (7.5,2) {$\vdots$};


\draw node at (8.5,2) {$\vdots$};

\draw (9, 4.5) rectangle (11,-.5);
\draw node at (10, 2) {$\mathbf{C}(\bm x, \bm \theta)$};

\end{tikzpicture}}
\caption{ Generic architecture of Variational Quantum Algorithm (VQA). The block  $\mathbf{U}(\bm x)$ denotes the data encoding circuit, where $\bm x$ is the input data. This is followed by the parameterized quantum circuits of variational circuit block $\mathbf{V}(\bm \theta)$, which consists of trainable parameters $\bm \theta$. After, a quantum measurement operation is performed on all qubits. Finally, the cost function $\mathbf{C}(\bm x, \bm \theta)$ is evaluated.}
\label{fig:variational quantum algorithm}
\end{center}
\end{figure*}

Variational Quantum Algorithms (VQAs) are hybrid quantum-classical frameworks that leverage quantum properties such as superposition and entanglement to enhance the efficiency of solving optimization tasks.
VQAs are considered parameterized quantum circuits or variational circuits, designed to train the circuit parameters iteratively according to the given optimization task.
A VQA typically consists of four core components: an encoding layer $\mathbf{U}(\bm{x})$, a parameterized layer $\mathbf{V}(\bm{\theta})$, a cost function $\mathbf{C}$, and an optimizing procedure to update the parameters $\bm{\theta}$. Figure \ref{fig:variational quantum algorithm} illustrates the generic architecture of a Variational Quantum Algorithm (VQA).

\begin{figure*}[ht]
\begin{center}
\resizebox{0.7\textwidth}{!}{
\begin{tikzpicture}
\tikzstyle{surround} = [fill=lime!10,thick,draw = black,rounded corners = 2mm]
\draw (0.3,0)--(1,0);
\draw (0.3,1)--(1,1);
\draw (0.3,2)--(1,2);
\draw (1, .4) rectangle (2.5,-.4);
\draw node at (1.75, 0) {$\mathbf R_z(v_1)$};
\draw (1, 1.4) rectangle (2.5,.6);
\draw node at (1.75, 1) {$\mathbf R_z(v_2)$};
\draw (1, 2.4) rectangle (2.5,1.6);
\draw node at (1.75, 2) {$\mathbf R_z(v_3)$};
\draw (2.5,0)--(3.2,0);
\draw (2.5,1)--(3.2,1);
\draw (2.5,2)--(3.2,2);
\draw node at (1.6, -1) {(a)};
\end{tikzpicture}

\begin{tikzpicture}
\tikzstyle{surround} = [fill=lime!10,thick,draw = black,rounded corners = 2mm]
\draw node at (3, 0){};
\draw node at (0, 0){};

\end{tikzpicture}

\begin{tikzpicture}
\tikzstyle{surround} = [fill=lime!10,thick,draw = black,rounded corners = 2mm]
\draw (0.3,0)--(.7,0);
\draw (0.3,1)--(.7,1);
\draw (0.3,2)--(.7,2);
\draw (.7, .4) rectangle (1.7,-.4);
\draw node at (1.2, 0) {$\mathbf  H$};
\draw (.7, 1.4) rectangle (1.7,.6);
\draw node at (1.2, 1) {$\mathbf H$};
\draw (.7, 2.4) rectangle (1.7,1.6);
\draw node at (1.2, 2) {$\mathbf H$};
\draw (1.7,0)--(2.1,0);
\draw (1.7,1)--(2.1,1);
\draw (1.7,2)--(2.1,2);
\draw (2.1, .4) rectangle (3.1,-.4);
\draw node at (2.6, 0) {$\mathbf P_3$};
\draw (2.1, 1.4) rectangle (3.1,.6);
\draw node at (2.6, 1) {$\mathbf P_2$};
\draw (2.1, 2.4) rectangle (3.1,1.6);
\draw node at (2.6, 2) {$\mathbf P_1$};
\draw (3.1,0)--(3.5,0);
\draw (3.1,1)--(3.5,1);
\draw (3.1,2)--(3.5,2);
\draw (3.5,2)--(3.5,.8);
\filldraw (3.5,2) circle (2pt);
\draw (3.5,1) circle (5pt);
\draw (3.5,0)--(6.1,0);
\draw (3.5,1)--(3.9,1);
\draw (3.5,2)--(8.9,2);
\draw (3.9, 1.4) rectangle (4.9,.6);
\draw node at (4.4, 1) {$\mathbf P_{1,2}$};

\draw (4.9,1)--(8.9,1);
\draw (5.3,2)--(5.3,.8);
\filldraw (5.3,2) circle (2pt);
\draw (5.3,1) circle (5pt);
\draw (8.9,1)--(11,1);
\draw (8.9,2)--(11,2);
\draw (5.7,2)--(5.7,-.2);
\filldraw (5.7,2) circle (2pt);
\draw (5.7,0) circle (5pt);
\draw (6.1, .4) rectangle (7.1,-.4);
\draw node at (6.6, 0) {$\mathbf P_{1,3}$};
\draw (7.1,0)--(7.5,0);
\draw (7.5,2)--(7.5,-.2);
\filldraw (7.5,2) circle (2pt);
\draw (7.5,0) circle (5pt);
\draw (7.5,0)--(8.7,0);

\draw (8.3,1)--(8.3,-.2);
\filldraw (8.3,1) circle (2pt);
\draw (8.3,0) circle (5pt);
\draw (8.7, .4) rectangle (9.7,-.4);
\draw node at (9.2, 0) {$\mathbf P_{2,3}$};
\draw (9.7,0)--(11,0);
\draw (10.1,1)--(10.1,-.2);
\filldraw (10.1,1) circle (2pt);
\draw (10.1,0) circle (5pt);
\draw node at (5.5, -1) {(b)};

\end{tikzpicture}}
\caption{(a) Angle Embedding. The feature vector is $\bm v = (v_1, v_2, v_3)$, encoded into 3 qubits. Rotation gates $\mathbf R_z$ are applied to encode the features; if not specified, $\mathbf R_x$ rotations are used by default. (b) ZZ Feature Map. The feature vector is $\bm v = (v_1, v_2, v_3)$, encoded into three qubits and one repetition Layer. $\mathbf{P}_i$ = $\mathbf{P}(2 * \psi(v_i))$ and  $\mathbf{P}_{i,j}$ = $\mathbf{P}(2 * \psi(v_i, v_j))$, where $\mathbf P$ denotes the Phase Gate $\mathbf P(\lambda) = \begin{pmatrix}
1 & 0 \\
0 & e^{i\lambda}
\end{pmatrix}$ and $\psi$ is a non-linear function, which defaults to $\psi(x) = x$, and $\psi(x, y) = (\pi - x)(\pi - y)$.}
\label{fig:AngleEmbedding and ZZ Feature Map}
\end{center}
\end{figure*}
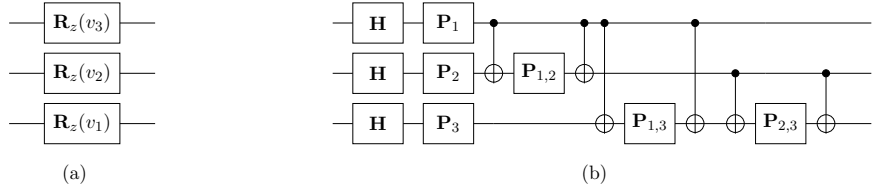

Classical information is first encoded into a quantum state via a state preparation routine or feature map~\cite{schuld2021effect}.
The choice of the feature map depends on the specified problem, as it significantly influences model performance and convergence speed. Notably, this feature map is neither trained nor optimized during training~\cite{lloyd2020quantumembeddingsmachinelearning}.
Here in Figure \ref{fig:AngleEmbedding and ZZ Feature Map}, we present two feature maps widely used in quantum machine learning: the AngleEmbedding~\cite{bergholm2018pennylane} and ZZFeatureMap~\cite{fingerhuth2018open}.

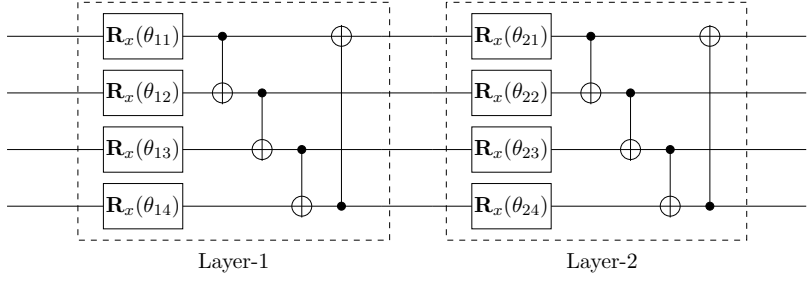
\begin{figure*}[ht]
\begin{center}
\resizebox{0.65\textwidth}{!}{
\begin{tikzpicture}
\tikzstyle{surround} = [fill=lime!10,thick,draw = black,rounded corners = 2mm]

\draw (1.5,0)--(3.2,0);
\draw (1.5,1)--(3.2,1);
\draw (1.5,2)--(3.2,2);
\draw (1.5,3)--(3.2,3);
\draw (3.2, .4) rectangle (4.6,-.4);
\draw node at (3.9, 0) {$\mathbf{R}_x(\theta_{14})$};
\draw (3.2, 1.4) rectangle (4.6,.6);
\draw node at (3.9, 1) {$\mathbf{R}_x(\theta_{13})$};
\draw (3.2, 2.4) rectangle (4.6,1.6);
\draw node at (3.9, 2) {$\mathbf{R}_x(\theta_{12})$};
\draw (3.2, 3.4) rectangle (4.6,2.6);
\draw node at (3.9, 3) {$\mathbf{R}_x(\theta_{11})$};
\draw (4.6,0)--(8.1,0);
\draw (4.6,1)--(8.1,1);
\draw (4.6,2)--(8.1,2);
\draw (4.6,3)--(8.1,3);
\draw (5.3,3)--(5.3,1.8);
\filldraw (5.3,3) circle (2pt);
\draw (5.3,2) circle (5pt);
\draw (6,2)--(6,.8);
\filldraw (6,2) circle (2pt);
\draw (6,1) circle (5pt);
\draw (6.7,1)--(6.7,-.2);
\filldraw (6.7,1) circle (2pt);
\draw (6.7,0) circle (5pt);
\draw (7.4,0)--(7.4,3.2);
\filldraw (7.4,0) circle (2pt);
\draw (7.4,3) circle (5pt);
\draw (8.1,0)--(9,0);
\draw (8.1,1)--(9,1);
\draw (8.1,2)--(9,2);
\draw (8.1,3)--(9,3);
\draw (9,0)--(9.7,0);
\draw (9,1)--(9.7,1);
\draw (9,2)--(9.7,2);
\draw (9,3)--(9.7,3);
\draw (9.7, .4) rectangle (11.1,-.4);
\draw node at (10.4, 0) {$\mathbf{R}_x(\theta_{24})$};
\draw (9.7, 1.4) rectangle (11.1,.6);
\draw node at (10.4, 1) {$\mathbf{R}_x(\theta_{23})$};
\draw (9.7, 2.4) rectangle (11.1,1.6);
\draw node at (10.4, 2) {$\mathbf{R}_x(\theta_{22})$};
\draw (9.7, 3.4) rectangle (11.1,2.6);
\draw node at (10.4, 3) {$\mathbf{R}_x(\theta_{21})$};
\draw (11.1,0)--(15.6,0);
\draw (11.1,1)--(15.6,1);
\draw (11.1,2)--(15.6,2);
\draw (11.1,3)--(15.6,3);
\draw (11.8,3)--(11.8,1.8);
\filldraw (11.8,3) circle (2pt);
\draw (11.8,2) circle (5pt);
\draw (12.5,2)--(12.5,.8);
\filldraw (12.5,2) circle (2pt);
\draw (12.5,1) circle (5pt);
\draw (13.2,1)--(13.2,-.2);
\filldraw (13.2,1) circle (2pt);
\draw (13.2,0) circle (5pt);
\draw (13.9,0)--(13.9,3.2);
\filldraw (13.9,0) circle (2pt);
\draw (13.9,3) circle (5pt);
\draw [dashed] (2.75, 3.6) rectangle (8.25, -.6);
\draw node at (5.5,-1){   Layer-1};
\draw [dashed] (9.25, 3.6) rectangle (14.55 , -.6);
\draw node at (12,-1){  Layer-2};
\end{tikzpicture}} 
\caption{Diagram of Basic Entangler Layers. Each layer (dashed box) comprises one-parameter single-qubit rotations on each qubit, followed by a closed chain of CNOT gates entangling consecutive qubits.}
\label{fig:Basic Entangler Layers}
\end{center}
\end{figure*}

\begin{figure*}[ht]
\begin{center}
\resizebox{0.65\textwidth}{!}{
\begin{tikzpicture}
\tikzstyle{surround} = [fill=lime!10,thick,draw = black,rounded corners = 2mm]
\draw (0.1,0)--(.7,0);
\draw (0.1,1)--(.7,1);
\draw (0.1,2)--(.7,2);
\draw (0.1,3)--(.7,3);
\draw (0.1,4)--(.7,4);
\draw (.7, .4) rectangle (2.1,-.4);
\draw node at (1.4, 0) {$\mathbf{R}_x(\theta_{15})$};
\draw (.7, 1.4) rectangle (2.1,.6);
\draw node at (1.4, 1) {$\mathbf{R}_x(\theta_{14})$};
\draw (.7, 2.4) rectangle (2.1,1.6);
\draw node at (1.4, 2) {$\mathbf{R}_x(\theta_{13})$};
\draw (.7, 3.4) rectangle (2.1,2.6);
\draw node at (1.4, 3) {$\mathbf{R}_x(\theta_{12})$};
\draw (.7, 4.4) rectangle (2.1,3.6);
\draw node at (1.4, 4) {$\mathbf{R}_x(\theta_{11})$};
\draw (2.1,0)--(7,0);
\draw (2.1,1)--(2.5,1);
\draw (2.1,2)--(7,2);
\draw (2.1,3)--(2.5,3);
\draw (2.1,4)--(7,4);
\draw (2.5, 3.4) rectangle (3.9,2.6);
\draw node at (3.2, 3) {$\mathbf{R}_z(\theta_{16})$};
\draw (3.2,4)--(3.2,3.4);
\filldraw (3.2,4) circle (2pt);
\draw (2.5, 1.4) rectangle (3.9,.6);
\draw node at (3.2, 1) {$\mathbf{R}_z(\theta_{17})$};
\draw (3.2,2)--(3.2,1.4);
\filldraw (3.2,2) circle (2pt);
\draw (3.9,3)--(7,3);
\draw (3.9,1)--(7,1);
\draw (4.3,4)--(4.3,1.8);
\filldraw (4.3,4) circle (2pt);
\filldraw (4.3,3) circle (2pt);
\draw (4.3,2) circle (5pt);
\draw (4.7,4)--(4.7,.8);
\filldraw (4.7,4) circle (2pt);
\filldraw (4.7,2) circle (2pt);
\draw (4.7,1) circle (5pt);
\draw (5.1,0)--(5.1, 3.2);
\filldraw (5.1,0) circle (2pt);
\filldraw (5.1,2) circle (2pt);
\draw (5.1,3) circle (5pt);
\draw (5.5,1)--(5.5, 4.2);
\filldraw (5.5,1) circle (2pt);
\filldraw (5.5,3) circle (2pt);
\draw (5.5,4) circle (5pt);

\draw (7, .4) rectangle (8.4,-.4);
\draw node at (7.7, 0) {$\mathbf{R}_x(\theta_{25})$};
\draw (7, 1.4) rectangle (8.4,.6);
\draw node at (7.7, 1) {$\mathbf{R}_x(\theta_{24})$};
\draw (7, 2.4) rectangle (8.4,1.6);
\draw node at (7.7, 2) {$\mathbf{R}_x(\theta_{23})$};
\draw (7, 3.4) rectangle (8.4,2.6);
\draw node at (7.7, 3) {$\mathbf{R}_x(\theta_{22})$};
\draw (7, 4.4) rectangle (8.4,3.6);
\draw node at (7.7, 4) {$\mathbf{R}_x(\theta_{21})$};
\draw (8.4,0)--(12.5,0);
\draw (8.4,1)--(8.8,1);
\draw (8.4,2)--(12.5,2);
\draw (8.4,3)--(8.8,3);
\draw (8.4,4)--(12.5,4);
\draw (8.8, 3.4) rectangle (10.2,2.6);
\draw node at (9.5, 3) {$\mathbf{R}_z(\theta_{26})$};
\draw (9.5,4)--(9.5,3.4);
\filldraw (9.5,4) circle (2pt);
\draw (8.8, 1.4) rectangle (10.2,.6);
\draw node at (9.5, 1) {$\mathbf{R}_z(\theta_{27})$};
\draw (9.5,2)--(9.5,1.4);
\filldraw (9.5,2) circle (2pt);
\draw (10.2,3)--(12.5,3);
\draw (10.2,1)--(12.5,1);
\draw (10.6,4)--(10.6,1.8);
\filldraw (10.6,4) circle (2pt);
\filldraw (10.6,3) circle (2pt);
\draw (10.6,2) circle (5pt);
\draw (11,4)--(11,.8);
\filldraw (11,4) circle (2pt);
\filldraw (11,2) circle (2pt);
\draw (11,1) circle (5pt);
\draw (11.4,0)--(11.4, 3.2);
\filldraw (11.4,0) circle (2pt);
\filldraw (11.4,2) circle (2pt);
\draw (11.4,3) circle (5pt);
\draw (11.8,1)--(11.8, 4.2);
\filldraw (11.8,1) circle (2pt);
\filldraw (11.8,3) circle (2pt);
\draw (11.8,4) circle (5pt);

\draw (12.5, .4) rectangle (13.9,-.4);
\draw node at (13.2, 0) {$\mathbf{R}_x(\theta_{35})$};
\draw (12.5, 1.4) rectangle (13.9,.6);
\draw node at (13.2, 1) {$\mathbf{R}_x(\theta_{34})$};
\draw (12.5, 2.4) rectangle (13.9,1.6);
\draw node at (13.2, 2) {$\mathbf{R}_x(\theta_{33})$};
\draw (12.5, 3.4) rectangle (13.9,2.6);
\draw node at (13.2, 3) {$\mathbf{R}_x(\theta_{32})$};
\draw (12.5, 4.4) rectangle (13.9,3.6);
\draw node at (13.2, 4) {$\mathbf{R}_x(\theta_{31})$};
\draw (13.9,0)--(16.1,0);
\draw (13.9,1)--(14.3,1);
\draw (13.9,2)--(16.1,2);
\draw (13.9,3)--(14.3,3);
\draw (13.9,4)--(16.1,4);
\draw (14.3, 3.4) rectangle (15.7,2.6);
\draw node at (15, 3) {$\mathbf{R}_z(\theta_{36})$};
\draw (15,4)--(15,3.4);
\filldraw (15,4) circle (2pt);
\draw (14.3, 1.4) rectangle (15.7,.6);
\draw node at (15, 1) {$\mathbf{R}_z(\theta_{37})$};
\draw (15,2)--(15,1.4);
\filldraw (15,2) circle (2pt);
\draw (15.7,3)--(16.1,3);
\draw (15.7,1)--(16.1,1);

\draw [dashed] (0.45, 4.6) rectangle (5.9, -.6);
\draw [dashed] (6.75, 4.6) rectangle (12.25, -.6);
\draw node at (3.25,-1){   Layer-1};
\draw node at (9.5,-1){  Layer-2};

\end{tikzpicture}} 
\caption{Diagram of N-local circuits. Each layer (dashed box) consists of Rotation blocks formed by $\mathbf R_x$ and 
C$\mathbf R_Z$ gates, followed by entanglement blocks formed by Toffoli gates. At the end of all layers, there is a Rotational block without entanglement.}
\label{fig:N-local circuits}
\end{center}
\end{figure*}
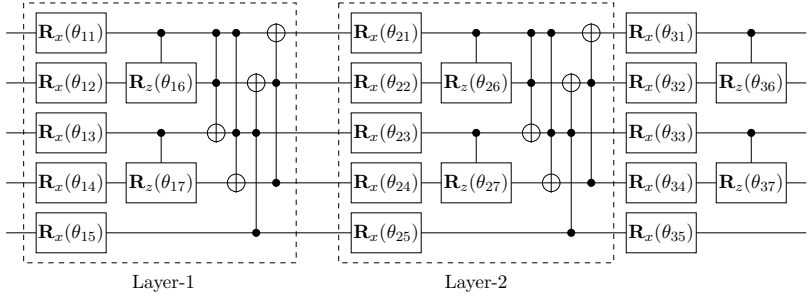

Once the classical data is encoded in the quantum device, a parametrized circuit~\cite{mitarai2018quantum, benedetti2019parameterized, schuld2020circuit} is applied to it. 
The parametrized circuit is the main component of VQAs, enabling them to learn and adapt during the optimization iteration.
A parametrized circuit consists of the quantum gates - such as $\mathbf R_x, \mathbf R_y, \mathbf R_z$ - whose parameters are learnable during iterations. These gates, when combined with quantum phenomena like superposition and entanglement between qubits, enable the circuit to capture complex model functions and optimize performance over successive iterations.
Figure~\ref {fig:Basic Entangler Layers} and Figure~\ref{fig:N-local circuits} are two examples of quantum parametrized circuits - Basic Entangler layers~\cite{bergholm2018pennylane} and N-local circuit~\cite{fingerhuth2018open} - commonly used in several variational quantum algorithms. 

After the quantum parametrized circuit, classical information is extracted from the quantum circuit through a quantum measurement operation on a subset (or all) of the circuit's qubits.
Measurement is an important framework of a quantum system, and Qiskit provides two primitives that can help to measure: Sampler and Estimator~\cite{fingerhuth2018open}.
The sampler primitives calculate the probability of a quantum state with respect to each computational basis state. Let there be a quantum circuit that prepares a quantum state $\ket{\psi}$. Then, sampler primitives calculate
\begin{align*}
    \text{P}_k = |\braket{k|\psi}|^2,
\end{align*}
here $\text{P}_k$ denotes the probability of measuring quantum state $\ket{\psi}$ with respect to the computational quantum state $\ket{k}$.
The estimator primitives introduce a different notion called the observable $\mathbf{\tilde{H}}$, which is a Hermitian linear operator.
Estimator primitives calculate the expectation value of $\mathbf{\tilde{H}}$ with respect to a given quantum state. 
Let $\ket{\lambda}$ be one of the eigenvector of the observable $\mathbf{\tilde{H}}$ with corresponding eigenvalue $\lambda$, then the observable probabilities are determined as: $\text{P}_\lambda = |\braket{\lambda|\psi}|^2$. 
The expectation value of the observable $\mathbf{\tilde{H}}$ with respect to a quantum state $\ket{\psi}$ is defined as the weighted sum of its eigenvalues $\lambda$, where each weight corresponds to the observable probability $\text{P}_\lambda$, i,e.,
\begin{align*}
    \braket{\mathbf{\tilde{H}}}_{\psi} = \braket{\psi|\mathbf{\tilde{H}}|\psi} = \sum_{\lambda}\text{P}_\lambda\lambda.
\end{align*}
The outcomes of this measurement are then fed into a cost function, defined by the optimization model. This cost function evaluates the performance of the parameterized quantum circuit and guides the update of its parameters during training.
Based on the cost function, an optimization algorithm - either gradient-based or gradient-free is applied to minimize or maximize the objective.
This process updates the parameters of the quantum circuit, which is then executed iteratively until convergence or for a fixed number of epochs.
After completing the iterative steps, the quantum circuit is considered optimized for the given model and produces an approximate optimal solution.

One of the most important advantages of VQA is its robustness against quantum noise~\cite{kandala2017hardware, farhi2014quantum, mcclean2016theory}, making it suitable for implementation on today’s Noisy Intermediate-Scale Quantum (NISQ) devices.
VQAs have been successfully applied across various domains in machine learning and artificial intelligence, including classification~\cite{schuld2020circuit, havlivcek2019supervised, farhi2018classification, benedetti2019parameterized}, generative modeling~\cite{dallaire2018quantum}, deep reinforcement learning~\cite{chen2020variational}, and transfer learning~\cite{mari2020transfer}.

\section{Quantum Temporal Fusion Transformers}
In this paper, we extend the classical Temporal Fusion Transformer (TFT) model into the Quantum Temporal Fusion Transformer (QTFT) model by replacing and appropriately modifying classical learning components within the TFT cell with VQCs.
There are three main components in TFT responsible for extracting the pattern from the datasets: Gated Residual Networks (GRNs), Long Short Term Memory (LSTM), and Interpretable Multi-head Attention Mechanism.
In this section, we focus on two key components: Gated Residual Network and  Interpretable Multi-head Attention Mechanisms, including all their associated sub-component. 
We are not focused on Long Short Term Memory in this work, as it has already been introduced~\cite{chen2022quantum}. 

\subsection{Variational Quantum Circuit for QTFT}
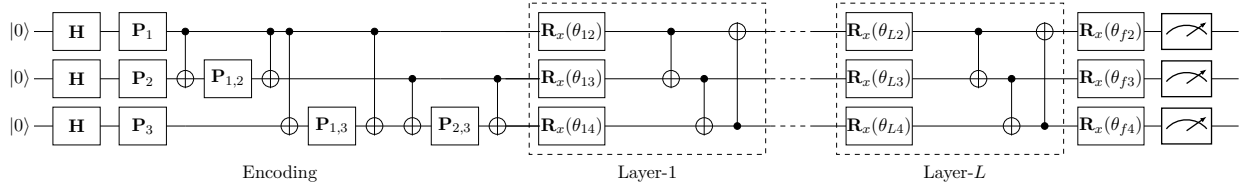
\begin{figure*}[ht]
\begin{center}
\resizebox{\textwidth}{!}{
\begin{tikzpicture}
\tikzstyle{surround} = [fill=lime!10,thick,draw = black,rounded corners = 2mm]
\draw node at (0, 0) {$\ket{0}$};
\draw node at (0, 1) {$\ket{0}$};
\draw node at (0, 2) {$\ket{0}$};
\draw (0.3,0)--(.7,0);
\draw (0.3,1)--(.7,1);
\draw (0.3,2)--(.7,2);
\draw (.7, .4) rectangle (1.7,-.4);
\draw node at (1.2, 0) {$\mathbf  H$};
\draw (.7, 1.4) rectangle (1.7,.6);
\draw node at (1.2, 1) {$\mathbf H$};
\draw (.7, 2.4) rectangle (1.7,1.6);
\draw node at (1.2, 2) {$\mathbf H$};
\draw (1.7,0)--(2.1,0);
\draw (1.7,1)--(2.1,1);
\draw (1.7,2)--(2.1,2);
\draw (2.1, .4) rectangle (3.1,-.4);
\draw node at (2.6, 0) {$\mathbf P_3$};
\draw (2.1, 1.4) rectangle (3.1,.6);
\draw node at (2.6, 1) {$\mathbf P_2$};
\draw (2.1, 2.4) rectangle (3.1,1.6);
\draw node at (2.6, 2) {$\mathbf P_1$};
\draw (3.1,0)--(3.5,0);
\draw (3.1,1)--(3.5,1);
\draw (3.1,2)--(3.5,2);
\draw (3.5,2)--(3.5,.8);
\filldraw (3.5,2) circle (2pt);
\draw (3.5,1) circle (5pt);
\draw (3.5,0)--(6.1,0);
\draw (3.5,1)--(3.9,1);
\draw (3.5,2)--(8.9,2);
\draw (3.9, 1.4) rectangle (4.9,.6);
\draw node at (4.4, 1) {$\mathbf P_{1,2}$};

\draw (4.9,1)--(8.9,1);
\draw (5.3,2)--(5.3,.8);
\filldraw (5.3,2) circle (2pt);
\draw (5.3,1) circle (5pt);
\draw (8.9,1)--(11,1);
\draw (8.9,2)--(11,2);
\draw (5.7,2)--(5.7,-.2);
\filldraw (5.7,2) circle (2pt);
\draw (5.7,0) circle (5pt);
\draw (6.1, .4) rectangle (7.1,-.4);
\draw node at (6.6, 0) {$\mathbf P_{1,3}$};
\draw (7.1,0)--(7.5,0);
\draw (7.5,2)--(7.5,-.2);
\filldraw (7.5,2) circle (2pt);
\draw (7.5,0) circle (5pt);
\draw (7.5,0)--(8.7,0);

\draw (8.3,1)--(8.3,-.2);
\filldraw (8.3,1) circle (2pt);
\draw (8.3,0) circle (5pt);
\draw (8.7, .4) rectangle (9.7,-.4);
\draw node at (9.2, 0) {$\mathbf P_{2,3}$};
\draw (9.7,0)--(11,0);
\draw (10.1,1)--(10.1,-.2);
\filldraw (10.1,1) circle (2pt);
\draw (10.1,0) circle (5pt);
\draw node at (5.5,-1){Encoding};
\end{tikzpicture}

\hspace{-1 cm}

\begin{tikzpicture}
\tikzstyle{surround} = [fill=lime!10,thick,draw = black,rounded corners = 2mm]
\draw (2.5,0)--(3.2,0);
\draw (2.5,1)--(3.2,1);
\draw (2.5,2)--(3.2,2);

\draw (3.2, .4) rectangle (4.6,-.4);
\draw node at (3.9, 0) {$\mathbf{R}_x(\theta_{14})$};
\draw (3.2, 1.4) rectangle (4.6,.6);
\draw node at (3.9, 1) {$\mathbf{R}_x(\theta_{13})$};
\draw (3.2, 2.4) rectangle (4.6,1.6);
\draw node at (3.9, 2) {$\mathbf{R}_x(\theta_{12})$};
\draw (4.6,0)--(8.1,0);
\draw (4.6,1)--(8.1,1);
\draw (4.6,2)--(8.1,2);
\draw (6,2)--(6,.8);
\filldraw (6,2) circle (2pt);
\draw (6,1) circle (5pt);
\draw (6.7,1)--(6.7,-.2);
\filldraw (6.7,1) circle (2pt);
\draw (6.7,0) circle (5pt);
\draw (7.4,0)--(7.4,2.2);
\filldraw (7.4,0) circle (2pt);
\draw (7.4,2) circle (5pt);
\draw [dashed](8.1,0)--(9,0);
\draw [dashed](8.1,1)--(9,1);
\draw [dashed](8.1,2)--(9,2);
\draw (9,0)--(9.7,0);
\draw (9,1)--(9.7,1);
\draw (9,2)--(9.7,2);
\draw (9.7, .4) rectangle (11.1,-.4);
\draw node at (10.4, 0) {$\mathbf{R}_x(\theta_{L4})$};
\draw (9.7, 1.4) rectangle (11.1,.6);
\draw node at (10.4, 1) {$\mathbf{R}_x(\theta_{L3})$};
\draw (9.7, 2.4) rectangle (11.1,1.6);
\draw node at (10.4, 2) {$\mathbf{R}_x(\theta_{L2})$};
\draw (11.1,0)--(14.6,0);
\draw (11.1,1)--(14.6,1);
\draw (11.1,2)--(14.6,2);
\draw (12.5,2)--(12.5,.8);
\filldraw (12.5,2) circle (2pt);
\draw (12.5,1) circle (5pt);
\draw (13.2,1)--(13.2,-.2);
\filldraw (13.2,1) circle (2pt);
\draw (13.2,0) circle (5pt);
\draw (13.9,0)--(13.9,2.2);
\filldraw (13.9,0) circle (2pt);
\draw (13.9,2) circle (5pt);

\draw (16,0)--(18,0);
\draw (16,1)--(18,1);
\draw (16,2)--(18,2);

\draw (14.6, .4) rectangle (16,-.4);
\draw node at (15.3, 0) {$\mathbf{R}_x(\theta_{f4})$};
\draw (14.6, 1.4) rectangle (16,.6);
\draw node at (15.3, 1) {$\mathbf{R}_x(\theta_{f3})$};
\draw (14.6, 2.4) rectangle (16,1.6);
\draw node at (15.3, 2) {$\mathbf{R}_x(\theta_{f2})$};

\node[meter, minimum width=30, minimum height=18, inner sep=7] (meter) at (16.9,0) {};
\node[meter, minimum width=30, minimum height=18, inner sep=7] (meter) at (16.9,1) {};
\node[meter, minimum width=30, minimum height=18, inner sep=7] (meter) at (16.9,2) {};

\draw [dashed] (3, 2.6) rectangle (8, -.6);
\draw node at (5.5,-1){  Layer-1};
\draw [dashed] (9.5, 2.6) rectangle (14.3, -.6);
\draw node at (12,-1){ Layer-$L$};
\end{tikzpicture}}
\caption{VQA architecture for the QTFT model. It consists of three layers: the data encoding layer, variational circuit layers (dashed boxes), and the quantum measurement layer (meter symbol).
Now, the number of qubits and measurements depends on the problem of interest. Also, the variational circuit, the dashed boxes, can be adopted according to the accuracy of the result by increasing the number of layers of the circuit, enabling the mode to capture more complex patterns effectively.}
\label{fig:VQC architecture for QTFT}
\end{center}
\end{figure*}

In this section, we build a variational quantum circuit that is used within the learning components of Gated Residual Networks (GRNs) and Interpretable Multi-head Attention Mechanism. 
See Figure~\ref{fig:VQC architecture for QTFT} for a schematic diagram of the Variational Quantum Circuit for QTFT.

There are various quantum simulator software platforms, such as PennyLane~\cite{bergholm2018pennylane} and IBM Qiskit~\cite{fingerhuth2018open}, that allow for calculating numerical evaluation of the quantum circuit on a classical computer.
In contrast, real quantum computers estimate these values through statistical sampling obtained from iterative measurements. 

\subsubsection{Encoding Layer}
Before performing any quantum computation within a quantum circuit, it is important to encode classical data into quantum states. This is achieved through an Encoding layer, the predefined technique or method to encode the classical data into the corresponding quantum state.    
Let $n$ be the number of qubits in a quantum system. Then, any quantum state $\ket{\phi}$ can be expressed as
\begin{equation*}
    \ket{\phi} = \sum_{i=0}^{2^n-1}\alpha_i\ket{i},
\end{equation*} 
where $\alpha_{i} \in \mathbb{C}$ represents the complex amplitudes associated with the computational basis state $\ket{i}$, where the index $i$ denotes the decimal representation of the bit-string.
The square of the amplitude $\alpha_{i}$ is the probability of measuring the quantum state in the basis state $\ket{i}$. These amplitudes must satisfy the normalization condition
\begin{equation*}
    \sum_{i=0}^{2^n-1}|\alpha_i|^2 = 1.
\end{equation*}
Encoding layers implement a systematic method to embed a classical vector $\bm v = (v_1, v_2, \dots, v_n)$ into a quantum state by mapping its coordinate values $v_j's$ to the amplitudes $\alpha_i's$ corresponding to a quantum state $\ket{\phi}$.

Here, we use the ZZ Feature Map, an encoding scheme in which a classical input vector transforms into a quantum state.
In the paper~\cite{havlivcek2019supervised}, the authors Havlíček et al. introduce the fundamental concept of the ZZ Feature Map. The circuit corresponding to the encoding technique is defined by the following unitary operator
\[
U(\bm v) = \exp\left(
i \sum_{j=1}^n v_j Z_j 
+ i \sum_{j < k} \psi(v_j, v_k) Z_j Z_k
\right),
\]
where $\psi$ be an non-liner function and  Pauli-$Z_j$ denoted as Pauli-$Z$ operator on the $j$-th qubit. The first term applies Z rotations encoding the features linearly as $\exp(i v_j Z_j)$, while the second term applies ZZ entangling rotations as $\exp(i \psi(v_j, v_k) Z_j Z_k)$.
Below, we describe a specific variant of the ZZ Feature Map. 

The first step is to create an equal superposition of all basis states from the initial state $\ket{0}^{\otimes n}$ using the Hadamard gate 
\begin{equation*}
    H(\ket{0}^{\otimes n}) = \frac{1}{\sqrt{2^n}}\sum_{i = 0}^{2^n -1}\ket{i}.
\end{equation*}
There are two major components in the ZZ Feature Map : 
a phase gate $\mathbf{P}$, define as 
\begin{equation*}
\mathbf P(\lambda) = \begin{pmatrix}
1 & 0 \\
0 & e^{i\lambda}
\end{pmatrix},
\end{equation*}
where $\lambda \in \mathbb{R}$ called rotation angle, and
a classical non-liner function $\psi$, which typically defaults to $\psi(x) = x$ for single-variable inputs and $\psi(x, y) = (\pi - x)(\pi - y)$ for pairwise interactions.
Each qubit $j$, after the application of the Hadamard gate, is transformed by a phase gate with a rotational angle $2 * \psi (v_j)$, where $v_j$ is the $j$-th component of the input vector $\bm v$.

We present a quantum routine that is repeatedly applied within the ZZ Feature Map. Let $v_i$ and $v_j$ denote the $i$-th and the $j$-th component of the input vector $\bm v$. 
For each such pair $(v_i, v_j)$, the routine applies the following sequence of quantum operations: two CNOT gates with target qubits $j$ and control qubit $i$, and between the two CNOT gates applies a phase gate with an angle $2 * \psi (v_i, v_j)$ to the $j$-th qubit. Here is the Figure.

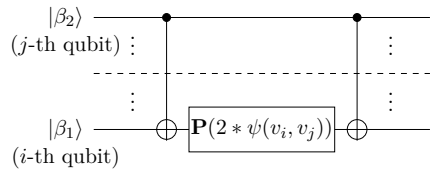
\begin{figure}[h!]
\begin{center}
\resizebox{0.35\textwidth}{!}{
\begin{tikzpicture}
\tikzstyle{surround} = [fill=lime!10,thick,draw = black,rounded corners = 2mm]
\draw node at (1.5,0) {$\ket{\beta_1}$};
\draw node at (1.5,2) {$\ket{\beta_2}$};
\draw node at (1.5,-.5) {($i$-th qubit)};
\draw node at (1.5,1.5) {($j$-th qubit)};
\draw (2,0)--(3.7, 0);
\draw (6.3, 0)--(8, 0);
\draw node at (2.7,.6) {$\vdots$};
\draw node at (7.3,.6) {$\vdots$};
\draw [dashed](2,1)--(8, 1);
\draw node at (2.7,1.6) {$\vdots$};
\draw node at (7.3,1.6) {$\vdots$};
\draw (2,2)--(8, 2);
\draw (3.3,-0.2)--(3.3,2);
\filldraw (3.3,2) circle (2pt);
\draw (3.3,0) circle (5pt);

\draw (3.7, .4) rectangle (6.3,-.4);
\draw node at (5, 0) {$\mathbf P(2 * \psi(v_i, v_j))$};

\draw (6.7,-0.2)--(6.7,2);
\filldraw (6.7,2) circle (2pt);
\draw (6.7,0) circle (5pt);


\end{tikzpicture}}
\end{center}
\end{figure}

The outputs of phase gates are passed through the above quantum routine in the following sequential order:
$(v_1,v_2),\dots, (v_1,v_n), (v_2,v_3), \dots, (v_2,v_n),$ $\dots$ $\, (v_k,v_{k+1}),\dots, (v_k,v_n), \dots, (v_{n-1},v_n)$.

As noted in reference~\cite{schuld2019quantum, havlivcek2019supervised}, this ZZ Feature Map offers several key advantages that leverage the computational power of the variational circuit,  particularly in the context of machine learning tasks.
This feature map provides nonlinear data encoding by mapping, where 
the data is projected into a high-dimensional space. 
Its structure enables the exploration of a larger portion of the Hilbert space, allowing it to capture more complex relationships within the data. It also provides a better starting point for the variational layer.

\subsubsection{Variational Circuit Layer}
The encoded data, in terms of the quantum state, is passed through a series of quantum unitary operators called a variational circuit.
In this variational quantum algorithm setup, we employ N-local circuits~\cite{qiskit2024} as the variational circuit or ansatz.
The N-local quantum unitary operators consist of single-qubit rotation gates with controlled-NOT (CNOT) gates.  
Single-qubit rotation gates $\mathbf R_y$ are implemented to each qubit with the rotational angle parameters $\theta _{(.)}$. Rotational angles are not predetermined; instead, they are iteratively updated during the optimization process using the gradient descent method.
To generate multi-qubit entanglement, the outputs of rotation gates are passed through CNOT gates between two consecutive qubits in cycle order: (1, 2, \dots $n-1$, 1).
A combination of rotation gates and CNOT gates is referred to as a layer, denoted as a dashed box in Figure~\ref{fig:VQC architecture for QTFT}. The layers are formulated as
\begin{align*}
  \bigotimes_{i=1}^n \mathbf R_y(\theta_i) \prod_{(i,j)} \text{CNOT}(i,j).
\end{align*}
Depending on the problem's complexity, the layer repeats several times to increase the circuit parameters, effectively capturing the more complex pattern of the dataset.
At the end of all layers, a final rotation layer consisting of $\mathbf R_y$ gates is appended. 

However, repeating the layers of the variational circuit increases the depth of the quantum circuit, which in turn affects the complexity and resource requirements of the quantum hardware. 
According to the problem and the limitations of current quantum hardware, it is important to optimize the depth of the circuit to produce the best possible result.

N-local circuits, particularly those with greater parameterized multi-qubit blocks, introduce a more expressive variational class, allowing the circuit to express more complicated quantum states. This greater expressivity will notably decrease the number of layers needed to synthesize an approximation of a target state. Experiments have shown that a worldwide entangling ansatz, with both two and three qubit gates on fully connected hardware, converges faster with significantly fewer layers than the standard two-local (hardware-efficient) ansatz. This results in exponential increases in convergence speed and iteration efficiency for a fixed target accuracy~\cite{du2022quantum, ayoub2025high}.

\subsubsection{Measurement Layer}
At the end of the variational circuit, a quantum measurement layer is added to extract quantum information for further post-processing on a classical computer.
In our variational quantum setup, we use a fixed, hardware-efficient Pauli observable~\cite{li2024quantum} - the Pauli-Z operator - as the measurement tool. The variational circuit is measured by applying the Pauli-Z observable independently to each qubit. Specifically,  for $i$-the qubit (where $i$ = 1, 2, 3, \dots, $n$), the observable is given by
\begin{align*}
    Z_i = I^{\otimes (i-1)} \otimes Z \otimes I^{\otimes (n-i)}.
\end{align*}

Let the quantum state after the variational circuit layer be denoted as $\ket{\zeta}$. We now demonstrate the calculation of measurement value by applying the Pauli-Z observable on the 0-th qubit, while the approach for calculating measurements on the remaining qubits follows the same.
The quantum state $\ket{\zeta}$ can be expressed in computational basis as   
\begin{align*}
    \ket{\zeta} &= \sum_{i = 0}^{2^n-1} \gamma_i\ket{i}\\
                &= \ket{0}\left( \sum_{i = 0}^{2^{(n-1)}-1} \gamma_i \ket{i}\right) \\
                & + \ket{1}\left( \sum_{i = 0}^{2^{(n-1)}-1} \gamma_{(2^{(n-1)} + i)} \ket{i}\right),
\end{align*}
where $\gamma_i \in \mathbb{C}$ and $\sum_{i = 0}^{2^n-1} |\gamma_i|^2 = 1$. 
The eigenvalues and eigenvectors of the observable Pauli-Z are 1, -1, and $\ket{0}$, $\ket{1}$ respectively, 
i.e., $\lambda_1 = 1$, $\lambda_2 = -1$ and $\ket{\lambda_1} = \ket{0}$, $\ket{\lambda_2} = \ket{1}$.  The probability, $\text{P}_1$ and $\text{P}_{-1}$ of measuring the quantum state $\ket{0}$ and $\ket{1}$ are given by 
\begin{align*}
    \text{P}_1 = \left | \braket{0|\zeta}                \right |^2
               &= \left | \sum_{i = 0}^{2^{(n-1)}-1} \gamma_i \ket{i} \right |^2\\
               &=  \sum_{i = 0}^{2^{(n-1)}-1} \left |\gamma_i \right |^2,\\
    \text{P}_{-1} = \left | \braket{1|\zeta}                \right |^2
               &= \left | \sum_{i = 0}^{2^{(n-1)}-1} \gamma_{(2^{(n-1)} + i)} \ket{i} \right |^2\\
               &=  \sum_{i = 0}^{2^{(n-1)}-1} \left |\gamma_{(2^{(n-1)} + i)} \right |^2.
\end{align*}
Then the expectation value of the Pauli-Z observable corresponding to the 0-th qubit is 
\begin{align*}
    \braket{\zeta|Z_0|\zeta}
    &= \text{P}_{\lambda_1} \lambda_1 + \text{P}_{\lambda_2} \lambda_2\\
    &= \sum_{i = 0}^{2^{(n-1)}-1} \left |\gamma_i \right |^2  -  \sum_{i = 0}^{2^{(n-1)}-1} \left |\gamma_{(2^{(n-1)} + i)} \right |^2.
\end{align*}

\subsection{QTFT Components}
The primary object of the QTFT model efficiently transforms key subroutines of TFT into quantum counterparts that leverage quantum computational advantage. We discuss this transformation in detail below.

\subsubsection{Quantum Gated Residual Network}

In the classical part of Section~\ref{CLASSICAL TEMPORAL FUSION TRANSFORMER}, we have already discussed the significance of the Gated Residual Network (GRN) in detail. In this section, we will not revisit its structure; instead, we will only focus on how this structure is adapted into a quantum form to improve the model's performance.  
In the previous section, we explored how classical neural network components (dense layers) can be replaced or alternated by quantum counterparts using Variational Quantum Algorithms (VQAs). Here, we utilize VQAs as the foundational building block of a Quantum Gated Residual Network (QGRN).

Let $\bm a$ and $\bm c$ denote the primary input and optional context input, respectively, where $\bm c$ is derived from a quantum static covariate encoder.
Fast, both the primary input $\bm a$ and the optional input $\bm c$ are plugged into the ZZ Feature map (denoted as ZZFeatureMap) to encode the classical data into quantum states $\ket{\bm a}$, $\ket{\bm c}$ respectively 
\begin{align*}
    \ket{\bm a} &= \text{ZZFeatureMap}(\bm a),\\
    \ket{\bm c} &= \text{ZZFeatureMap}(\bm c). 
\end{align*}

Two quantum states $\ket{\bm a} \text{and} \ket{\bm c}$ are passed independently through two separate variational circuits known as N-local circuits (denoted as NLocal). These circuits consist of parametrized quantum gates, in which parameters (or weights) are trainable during learning iterations. The resulting quantum states are present as $\ket{\bm a'}$ and $\ket{\bm c'}$
\begin{align*}
    \ket{\bm a'} &= \text{NLocal}_{\bm a'}(\ket{\bm a}),\\
    \ket{\bm c'} &= \text{NLocal}_{\bm c'}(\ket{\bm c}), 
\end{align*}
where the subscript $\bm a'$ denotes the trainable parameters associated with this particular entangler variational circuit. 
At the end of the quantum circuits, quantum measurement operations are implemented on quantum states $\ket{\bm a'}$ and $\ket{\bm c'}$ to extract classical information. The measurement is done by computing the expectation values concerning the Pauli-Z observable on each qubit
\begin{align*}
    \bm a'' &= \bra{\bm a'}\mathbf{Z}\ket{\bm a'} = \text{expval(PauliZ($\ket{\bm a'})$)},\\
    \bm c'' &= \bra{\bm c'}\mathbf{Z}\ket{\bm c'} =\text{expval(PauliZ($\ket{\bm c'})$)},
\end{align*}
where $\text{expval(PauliZ($\ket{\bm k})$}$ denote the expectation values concerning the Pauli-Z observable corresponding qubit $\ket{\bm k}$.
The classical two outputs obtained from quantum measurement, $\bm a''$ and $\bm c''$, are first added, followed by the ELU activation function to introduce non-linearity
\begin{align*}
    \bm \eta_1 = \text{ELU}(\bm a'' + \bm c'').
\end{align*}
This activated vector $\bm \eta_1$ is then encoded into the quantum state back using the ZZ Feature map for subsequent quantum processing
\begin{equation*}
    \ket{\bm \eta_1} = \text{ZZFeatureMap}(\bm \eta_1).
\end{equation*}
Another variational quantum circuit, N-local circuits, is applied to the quantum state $\bm \eta_1$ without performing any intermediate measurement operations
\begin{align*}
    \ket{\bm \eta_2} &= \text{NLocal}_{\bm \eta_2}(\ket{\bm \eta_1}).
\end{align*}
We introduce Quantum Gated Linear Unit (QGLU), a quantum analog of the classical Gated Linear Unit (GLU).
Let $\bm \gamma$ be the input of Quantum Gated Linear Unit (QGLU). If $\bm \gamma$ is a classical vector, it is first encoded into a quantum state using the ZZ Feature Map. Otherwise, if the input is already in a quantum state, this encoding step is omitted. Let $\ket{\bm \gamma}$ denote the encoded quantum state corresponding to input $\bm \gamma$. Then it is passed through two distinct variational quantum circuits, both implemented using N-local circuits
\begin{align*}
    \ket{\bm \gamma} &= \text{ZZFeatureMap}(\bm \gamma),\\
    \ket{\bm \gamma'} &= \text{NLocal}_{\bm \gamma'}(\ket{\bm \gamma}),\\
    \ket{\bm \gamma''} &= \text{NLocal}_{\bm \gamma''}(\ket{\bm \gamma}). 
\end{align*}
Now quantum measurement operations are applies to the quantum states $\ket{\bm \gamma'}$ and $\ket{\bm \gamma''}$
\begin{align*}
    \bm \gamma' &= \bra{\bm \gamma'}\mathbf{Z}\ket{\bm \gamma'}= \text{expval(PauliZ($ \ket{\bm \gamma'}$)},\\
    \bm \gamma'' &= \bra{\bm \gamma''}\mathbf{Z}\ket{\bm \gamma''} = \text{expval(PauliZ($ \ket{\bm \gamma''}$)}. 
\end{align*}
The final output of the Quantum Gated Linear Unit (QGLU) is computed using an element-wise multiplication between one of the sigmoid-activated outputs and another
\begin{align*}
    \bm \gamma''' = \text{QGLU}(\ket{\bm \gamma}) =  \sigma(\bm \gamma') \odot \bm \gamma'' ,
\end{align*}
where $\sigma(.)$ denote as sigmoid activation function and $\odot$ is the Hadamard product.

Now we are back to Quantum Gated Residual Network (QGRN), the final output of QGRN is a residual connection between the output of the Quantum Gated Linear Unit (QGLU) and the primary input, followed by  layer normalization
\begin{align*}
   \text{QGRN}(\bm a, \bm c) =  \text{LayerNorm}(\bm a + \text{QGLU}(\ket{\bm \eta_2})).
\end{align*}

\subsubsection{Quantum Variable Selection Network And Quantum Static Covariate Encoders}
Variable selection network and static covariate encoders are built upon GRN.
Now we have already constructed QGRN in the previous section. If we replace GRN by QGRN in both the variation selection network and the static covariate encoders, we derive the corresponding quantum variation selection network and quantum static covariate encoders.

\subsubsection{Quantum Interpretable Multi-head Attention}
In Section~\ref{CLASSICAL TEMPORAL FUSION TRANSFORMER}, we have already discussed the attention mechanism and how its modified version, called interpretable multi-head attention, efficiently improves the performance of the model.
In this section, we do not go through all the details; we focus only on building the architecture of interpretable multi-head attention within a quantum framework. 
The key components of the attention model are the learning parameters derived from three matrices: the query, key, and value metrics. A major problem in the classical model is efficiently learning and managing these large-scale parameters. 
The VQAs provide a quantum approach that can handle such parameters more effectively, potentially reducing computational overhead and improving learning efficiency.  
Below, we describe an approach for integrating VQAs into interpretable multi-head attention.

Let $\bm S$ be the input of the attention mechanism in matrix form and $m_H$ represent the number of attention heads. Each classical input row is first encoded into a quantum state by the ZZ Feature map
\begin{align*}
    \ket{\bm S} = \text{ZZFeatureMap}(\bm S),
\end{align*}
where we denote $\ket{\bm S}$ as the quantum states generated corresponding to all input rows.
We implement quantum variation circuits using N-local circuits to construct the query, key, and value. From the input $\ket{\bm S}$, we construct $m_H$ number of distinct queries and keys and one value
\begin{align*}
   \ket{\bm Q^{(h)}} &= \text{NLocal}_{Q^{(h)}}(\ket{\bm S}),\\
    \ket{\bm K^{(h)}} &= \text{NLocal}_{K^{(h)}}(\ket{\bm S}),\\
    \ket{\bm V} &= \text{NLocal}_{V}(\ket{\bm S}),
\end{align*}
for $h = 1, 2, \dots, m_H$.
To extract classical information from the quantum states of queries, keys, and values, apply quantum measurement operations with the Pauli-Z observable
\begin{align*}
    \bm Q^{(h)} &= \bra{\bm Q^{(h)}}\mathbf{Z}\ket{\bm Q^{(h)}} = \text{expval(PauliZ($\ket{\bm Q^{(h)}}$)},\\
    \bm K^{(h)} &= \bra{\bm K^{(h)}}\mathbf{Z}\ket{\bm K^{(h)}} = \text{expval(PauliZ($\ket{\bm K^{(h)}}$)},\\
    \bm V &= \bra{\bm V}\mathbf{Z}\ket{\bm V} = \text{expval(PauliZ($\ket{\bm V}$))},
\end{align*}
for $h = 1, 2, \dots, m_H$.
From this point onward, the Quantum Interpretable Multi-Head Attention mechanism operates analogously to its classical counterpart. It shares the same value $\bm V$ across all heads  while employing distinct query $\bm{Q}^{(.)}$ and key $\bm{K}^{(.)}$ projections for each head. The final output is obtained through additive aggregation of the attention outputs from all heads

\begin{align*}
    \text{QuantumInterpretableMultiHead}(\bm{Q}, \bm{K}, \bm{V}) = \frac{1}{m_H} \sum_{h=1}^{m_H} \text{Attention}\left( \bm{Q}^{(h)}, \bm{K}^{(h)}, \bm{V}\right ).
\end{align*}

The attention mechanism employed is identical to that used in the classical attention model. 
In classical models, a final linear projection is typically applied at the output. However, in our approach, we omit this projection since the variational circuit already contains enough number of learnable parameters.  
  
\subsection{Quantum Model Architecture}
This section explicitly discusses the Quantum Temporal Fusion Transformer (QTFT) architecture compared to its classical counterpart.  
As in the classical model, the Quantum Temporal Fusion Transformer (QTFT) also processes three kinds of input: static inputs, past inputs, and prior known future inputs.  

First, the static input passes through Quantum Variable Selection Networks, followed by Quantum Static Covariate Encoder, which produces three context vectors. 
Past inputs and prior known future inputs are also processed through the Quantum Variable Selection Networks, guided by one context vector that derives from the Quantum Static Covariate Encoder.
\begin{figure*}[ht]
\begin{center}
\resizebox{.55\textwidth}{!}{
\begin{tikzpicture}
\tikzstyle{surround} = [fill=lime!10,thick,draw = black,rounded corners = 2mm]

\draw (0, 1) rectangle (3, 0);
\draw node at (1.5, 0.5) { \small \shortstack{Quantum Variable\\Selection Network}};
\draw node at (1.5, -.75) {  Static Inputs};
\draw[->] (2,-.55)--(2,-.1);

\draw (5, 1) rectangle (8, 0);
\draw node at (6.5, 0.5) {  \small \shortstack{Quantum Variable \\Selection Network}};
\draw node at (7, -.75) {  Past Inputs};
\draw[->] (7,-.55)--(7,-.1);

\draw (10, 1) rectangle (13, 0);
\draw node at (11.5, 0.5) { \small  \shortstack{Quantum Variable \\ Selection Network}};

\draw node at (12, -.75) {  Future Inputs};
\draw[->] (12,-.55)--(12,-.1);

\draw (0, 2.5) rectangle (3, 1.5);
\draw node at (1.5, 2) { \small \shortstack {Quantum Static \\Covariate Encoder}};
\draw[->] (1.5,1)--(1.5,1.4);

\draw (5, 2.5) rectangle (8, 1.5);
\draw node at (6.5, 2) {\small  \shortstack {LSTM /\\ QLSTM}};
\draw[->] (6.5,1)--(6.5,1.4);

\draw (10, 2.5) rectangle (13, 1.5);
\draw node at (11.5, 2) { \small \shortstack {LSTM/\\ QLSTM}};
\draw[->] (11.5,1)--(11.5,1.4);

\draw[->] (3, 1.75)--(4,1.75)--(4, -0.4)--(11, -0.4)--(11, -0.1);
\draw[->] (6, -0.4)--(6, -0.1);

\draw[->] (3, 2.25)--(4.9, 2.25);
\draw[->] (8, 2.25)--(9.9, 2.25);

\draw (5, 4) rectangle (8,3);
\draw node at (6.5, 3.5) {\small  \shortstack {Quantum Gate\\ Add + Norm}};
\draw (10, 4) rectangle (13,3);
\draw node at (11.5, 3.5) {\small  \shortstack {Quantum Gate\\ Add + Norm}};
\draw[->] (6.5, 2.5)--(6.5, 2.9);
\draw[->] (11.5, 2.5)--(11.5, 2.9);
\draw[->] (6.5, 1.2)--(8.5, 1.2)--(8.5, 3.5)--(8, 3.5);
\draw[->] (11.5, 1.2)--(13.5, 1.2)--(13.5, 3.5)--(13.1, 3.5);
\draw (5, 4.5) rectangle (8,5);
\draw node at (6.5, 4.75) { \small {Quantum GRN} };
\draw (10, 4.5) rectangle (13,5);
\draw node at (11.5, 4.75){ \small {Quantum GRN} };
\draw[->] (7, 4)--(7, 4.4);
\draw[->] (12, 4)--(12, 4.4);

\draw[->] (1.5, 2.5)--(1.5, 4.2)--(11, 4.2)--(11, 4.4);
\draw[->] (6, 4.2)--(6, 4.4);
\draw (5, 5.5) rectangle (13,6);
\draw node at (9, 5.75) { \small {Quantum Interpretable Multi-Head Attention} };
\draw[->] (6.5, 5)--(6.5, 5.4);
\draw[->] (11.5, 5)--(11.5, 5.4);

\draw (10, 6.5) rectangle (13,7.5);
\draw node at (11.5, 7) { \small \shortstack {Quantum Gate\\ Add + Norm}};
\draw[->] (12, 4.2)--(13.75, 4.2)--(13.75, 9.5)--(13.1, 9.5);
\draw (10, 8) rectangle (13,8.5);
\draw node at (11.5, 8.25){ \small {Quantum GRN} };
\draw (10, 9) rectangle (13,10);
\draw node at (11.5, 9.5) {\small  \shortstack {Quantum Gate\\ Add + Norm}};
\draw[->] (11.5, 5.2)--(13.5, 5.2)--(13.5, 7)--(13.1, 7);
\draw[->] (11.5, 6)--(11.5, 6.4);
\draw[->] (11.5, 7.5)--(11.5, 7.9);
\draw[->] (11.5, 8.5)--(11.5, 8.9);
\draw (10, 10.5) rectangle (13,11);
\draw node at (11.5, 10.75){  {\small Dense}};
\draw[->] (11.5, 10)--(11.5, 10.4);
\draw[->] (11.5, 11)--(11.5, 11.5);
\draw (10, 11.2)--(13, 11.2 );
\draw[->] (10, 11.2)--(9.9, 11.4);
\draw[->] (13, 11.2)--(13.1, 11.4);
\draw node at (7.5, 11.7){\small  {Quantile forecasts :-}};

\draw node at (9.8, 11.7){\small  {$\hat{y}(0.1, t, \tau)$}};
\draw node at (11.5, 11.7){\small  {$\hat{y}(0.5, t, \tau)$}};
\draw node at (13.2, 11.7){\small  {$\hat{y}(0.9, t, \tau)$}};

\end{tikzpicture}}
\caption{QTFT architecture. QTFT processes three types of inputs: static inputs, time-dependent past inputs, and prior known future inputs. In this architecture, all classical components, including the variable selection network, static covariate encoder, gating layer, gated residual network, and interpretable multi-head attention, are systematically and efficiently transformed into quantum subroutines.}
\label{fig: QTFT}
\end{center}
\end{figure*}
The outputs of the Quantum Variable Selection Networks corresponding to the past inputs are passed through the LSTM Encoder, while those corresponding to the future inputs pass through the LSTM Decoder. The cell state and hidden state of the first LSTM in the layer are initialized using the context vector derived from the Quantum Static Covariate Encoder.
Rather than using a classical LSTM, we replace it with a Quantum Long Short-Term (QLSTM) memory~\cite{chen2022quantum}. However, to ensure a fair comparison between our proposed subroutines - Quantum GRN and Quantum Interpretable Multi-Head Attention - and their classical counterparts, we retain the classical LSTM as the base architecture.
The final outputs of this layer are obtained using Quantum Gated Linear Units (QGLUs) and applied through a residual connection followed by layer normalization.
Before applying Quantum Interpretable Multi-Head Attention, the output of Quantum Gated Layer Units is passed through a Quantum Gated Residual Network together with the last context vector from the  Quantum Static Covariate Encoder.
Both outputs of the Quantum Gated Residual Network, corresponding to past inputs and future inputs, are fed into the Quantum Interpretable Multi-Head Attention, followed by Quantum Gated Layer Units with residual connection and layer normalization.
The outputs of Quantum Gated Layer Units corresponding to future inputs are attached through a Quantum Gated Residual Network, followed by Quantum Gated Layer Units with residual connection and layer normalization.
Finally, quantile forecasts are obtained by applying dense layers to the outputs of Quantum Gate Layer Units.  

\subsection{Optimization Procedure}
The proposed architecture is a quantum circuit-based model, where each component is represented by a quantum circuit. In this section, we discuss an optimization technique for these quantum circuits to achieve the best possible result. 
Here, we use the gradient-based method to optimize the quantum circuits.  
Specifically, we utilize the parameter-shift rule~\cite{bergholm2018pennylane, schuld2019evaluating}, which enables the analytical computation of the gradient of the quantum circuits concerning their tunable parameters.
We are not going through the details of all the quantum circuit optimization procedures; instead, we illustrate a general quantum circuit optimization framework. The used quantum circuits in our architecture follow a similar structure, differing primarily in their inputs, variational circuits, and measurement configurations.

Let $\bm x$ denote the input data, $\mathbf{U(\bm x)}$ represent the data encoding unitary, and $\mathbf{V}(\bm \theta)$ be a variational circuit block, which consists of trainable parameters $\bm \theta$. Then the expectation value of an observable  $\tilde{\mathbf{H}}$ is given by
\begin{align*}
     \langle \tilde{\mathbf H} \rangle_{\bm x, \bm\theta} 
     &=  \bra{\mathbf{U^{\dagger}}(\bm x)\mathbf{V^{\dagger}}(\bm \theta)} \tilde{\mathbf H}\ket{\mathbf{V}(\bm \theta)\mathbf{U}(\bm x)}\\
     &= \bra{0}\mathbf{U^{\dagger}}(\bm x)\mathbf{V^{\dagger}}(\bm \theta)\tilde{\mathbf{H}}\mathbf{V}(\bm \theta)\mathbf{U}(\bm x)\ket{0}.
\end{align*}
It can be shown~\cite{mitarai2018quantum} that gradient of the function $\tilde{\mathbf{H}}$ with respect to $\bm \theta$ is given by
\begin{align*}
    \frac{\partial  \langle \tilde{\mathbf{H}} \rangle_{\bm x, \bm\theta}}{\partial \bm \theta}
     &= \frac{1}{2}\left [ \langle \tilde{\mathbf{H}} \rangle_{\bm x, \bm\theta + \frac{\pi}{2}}  - \langle\tilde{\mathbf{H}} \rangle_{\bm x, \bm\theta - \frac{\pi}{2}}    \right].
\end{align*}
Hence, it is proven that the gradient of the expectation values is evaluated analytically using the above equation. By combining this approach with classical gradient descent optimization, we obtained a quantum-based gradient descent optimization process and used it in our implementation.

\section{Experiments and Results}
This section presents a comparative analysis of the QTFT and its classical counterpart, focusing on their respective capabilities and performance.
Specifically, we present experimental results of multi-horizontal time series forecasting across various time series datasets by using the QTFT model.
We implemented the classical TFT model using the PyTorch~\cite{paszke2019pytorch} framework. For the simulation of quantum circuits in the QTFT model, we use PennyLane~\cite{bergholm2018pennylane}, while the overall architecture of QTFT is built using the framework PyTorch, as in the classical model.  
To ensure fair competition, we use the same structure for both the classical and QTFT models, including the cost function and fixed parameters.

Due to the limited number of available qubits and the inherent noise in current NISQ quantum devices, we do not use all the instances and features for training and testing our QTFT model. Also, we are not concerned with the data types, whether static, observed inputs, or known inputs, because we are working on a small data set. If we further divide the dataset into these categories, it would result in feature vectors that lack sufficient value to extract the relationships between the input variables.
Similarly, due to the limitations of current quantum hardware, we do not use the original loss function used in the TFT paper~\cite{lim2021temporal}. Instead, we use a simplified yet similar kind of loss function that efficiently calculates the loss for further optimization. Specifically, we employ the quantile loss function, defined as 
\[
\mathcal{L}_q(y, \hat{y}) = \frac{1}{m} \sum_{i=1}^m \max\left( (q - 1)(y_i - \hat{y}_i),\ q(y_i - \hat{y}_i) \right),
\]
where \( y_i \) denotes the true value of the \(i\)-th data point, \( \hat{y}_i \) represents the corresponding predicted value, \( q \in (0,1) \) specifies the target quantile for estimation, and \( m \) indicates the total number of data points in the dataset.

In Table~\ref{Table 1}, we present the fixed hyperparameters that are consistently used across the classical TFT and both variants of the QTFT model. These parameters include quantile, learning rate, number of epochs, input window (past steps), forecast steps (future steps), training data range, and test data range. 

\begin{table}[ht]
\centering
\begin{tabular}{|l|c|}
\hline
\textbf{Parameter} & \textbf{Value} \\ 
\hline
Quantile ($q$) & 0.5 \\ 
\hline
Learning Rate & 0.1 \\ 
\hline
Number of Epochs & 100 \\ 
\hline
Input Window (Past Steps) & 2 \\ 
\hline
Forecast Steps & 2 \\ 
\hline
Training Data Range & 0 -- 19 \\ 
\hline
Test Data Range & 20 -- 26 \\ 
\hline
\end{tabular}
\caption{Fixed Hyperparameters for Classical and Quantum TFT Models}
\label{Table 1}
\end{table}

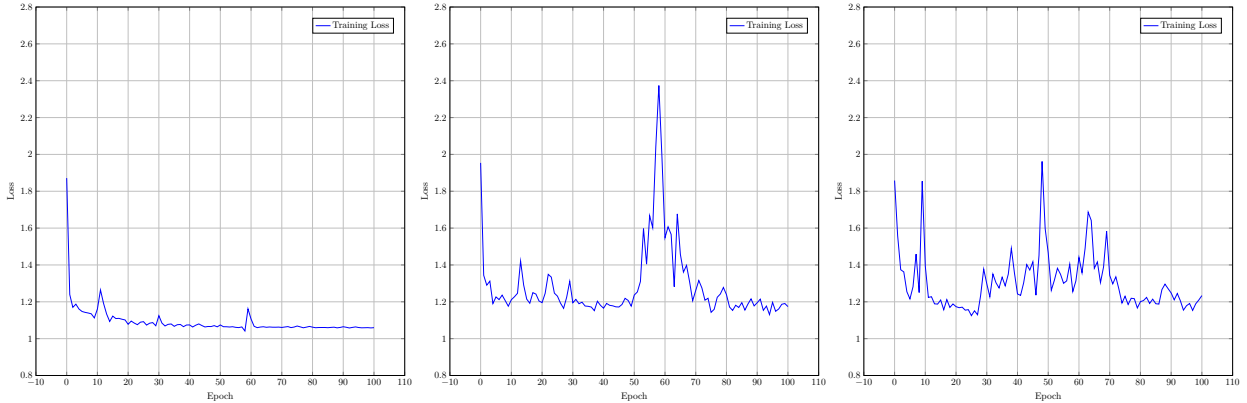
\begin{figure*}[ht]
\resizebox{\textwidth}{!}{
\begin{tikzpicture}
\centering
\begin{axis}[
    width=1\textwidth,
    height=1\textwidth,
    xlabel={Epoch},
    ylabel={Loss},
    grid=major,
    legend pos=north east,
    ymin= 0.8,
    ymax= 2.8
]
\addplot[
    color=blue,
    thick
] coordinates {
    (0,1.8714735731482506)(1,1.2395426221191883)(2,1.1700008250772953)(3,1.187098827213049)(4,1.1608527079224586)(5,1.14748015999794)(6,1.1426338590681553)(7,1.1392291709780693)(8,1.13481080904603)(9,1.1125871799886227)(10,1.1605703979730606)(11,1.264277521520853)(12,1.1913777068257332)(13,1.133429553359747)(14,1.0934453271329403)(15,1.1222984343767166)(16,1.1090456247329712)(17,1.110108207911253)(18,1.1053449660539627)(19,1.1015264727175236)(20,1.0779663436114788)(21,1.0953381732106209)(22,1.0845302939414978)(23,1.0758635513484478)(24,1.0893829427659512)(25,1.0921898037195206)(26,1.0736049562692642)(27,1.083921018987894)(28,1.0871049277484417)(29,1.0699513591825962)(30,1.1245454177260399)(31,1.0844695456326008)(32,1.0689283832907677)(33,1.0775705501437187)(34,1.07967109978199)(35,1.0668673068284988)(36,1.0756695829331875)(37,1.076931156218052)(38,1.065045315772295)(39,1.0741085931658745)(40,1.0745187178254128)(41,1.0632693134248257)(42,1.0729847438633442)(43,1.079853791743517)(44,1.0710760317742825)(45,1.063426062464714)(46,1.0660968571901321)(47,1.0656580924987793)(48,1.0706549249589443)(49,1.064173862338066)(50,1.0742848366498947)(51,1.0648054964840412)(52,1.0645457059144974)(53,1.063187338411808)(54,1.064692236483097)(55,1.0613427199423313)(56,1.059974018484354)(57,1.0635886825621128)(58,1.041536197066307)(59,1.1616244614124298)(60,1.1063713431358337)(61,1.066913578659296)(62,1.0597399100661278)(63,1.0629192665219307)(64,1.064594991505146)(65,1.0616720728576183)(66,1.0633235834538937)(67,1.0621668584644794)(68,1.0620385631918907)(69,1.062683466821909)(70,1.060833528637886)(71,1.0631958693265915)(72,1.0654436945915222)(73,1.059908077120781)(74,1.0629167072474957)(75,1.0685253851115704)(76,1.0639922060072422)(77,1.0589640736579895)(78,1.0624714605510235)(79,1.0663852617144585)(80,1.0623792298138142)(81,1.059300184249878)(82,1.0603537559509277)(83,1.060528364032507)(84,1.0605827271938324)(85,1.0594986788928509)(86,1.0610007494688034)(87,1.0622392073273659)(88,1.0586405619978905)(89,1.0608703941106796)(90,1.064639899879694)(91,1.0617037974298)(92,1.0579257383942604)(93,1.0610644295811653)(94,1.063645377755165)(95,1.0603942908346653)(96,1.058808945119381)(97,1.0592726916074753)(98,1.0597033314406872)(99,1.0582856498658657)(100,1.0596346259117126)

};
 
\legend{Training Loss}
\end{axis}
\end{tikzpicture}
\hfill
\begin{tikzpicture}
\begin{axis}[
    width=1\textwidth,
    height=1\textwidth,
    xlabel={Epoch},
    ylabel={Loss},
    grid=major,
    legend pos=north east,
    ymin=.8,
    ymax= 2.8
]
\addplot[
    color=blue,
    thick
] coordinates {

(0,1.9539776976884982)(1,1.343689194765293)(2,1.2902146326475863)(3,1.3121288533375635)(4,1.1905959246121314)(5,1.2264896939653007)(6,1.212118190147654)(7,1.2361089527387639)(8,1.2065351849052162)(9,1.1760792899555095)(10,1.211266055486747)(11,1.2272322693660231)(12,1.2461929895569401)(13,1.4213303924084744)(14,1.2911743412030912)(15,1.214412872515481)(16,1.192206338069089)(17,1.2496651337467117)(18,1.2418296560815107)(19,1.2043868657371082)(20,1.1956782878199816)(21,1.2460406116477492)(22,1.349005628220255)(23,1.3339238386564198)(24,1.2470044566095129)(25,1.2306472984140457)(26,1.192421400754434)(27,1.1647127943675377)(28,1.2247766682313233)(29,1.3108686727482868)(30,1.1949259976946696)(31,1.2131527321725488)(32,1.1896537308253392)(33,1.1976507753887438)(34,1.1770366902982698)(35,1.175542760591715)(36,1.171625280555683)(37,1.151903882905981)(38,1.20319647590665)(39,1.180814395657701)(40,1.1652551802211382)(41,1.191018866279277)(42,1.1808135841033787)(43,1.1781509494974876)(44,1.1726789146281438)(45,1.1718216580493785)(46,1.185403564624541)(47,1.2186809353913166)(48,1.2075462738685445)(49,1.176432831795781)(50,1.2366038252778084)(51,1.2533831623125495)(52,1.3100955393538973)(53,1.5973006057564445)(54,1.4043874261992093)(55,1.6671298827998353)(56,1.6023233736009455)(57,2.0397398016519976)(58,2.373866840663343)(59,1.9869116719859956)(60,1.5480954076977158)(61,1.6069523490099504)(62,1.5657743336656142)(63,1.2820451505004935)(64,1.6770600984936566)(65,1.4575203608501837)(66,1.3609664577637697)(67,1.3977761445179084)(68,1.3087249863183017)(69,1.2058947048355853)(70,1.261872970707358)(71,1.3154550246022505)(72,1.275265591063063)(73,1.2086499747740607)(74,1.2190849127227408)(75,1.142963369805205)(76,1.1591902834145364)(77,1.2245684804862556)(78,1.2420924112385734)(79,1.2778250268310907)(80,1.2388689078309358)(81,1.171427725891229)(82,1.1533043894163437)(83,1.1811617465011057)(84,1.1693010953369958)(85,1.1952194848952589)(86,1.1550497700489664)(87,1.1882284785633934)(88,1.2159080891608078)(89,1.1777471283759349)(90,1.194907708000053)(91,1.2143910393386805)(92,1.153887946626613)(93,1.1767628013512872)(94,1.1305537019656922)(95,1.1969151450297915)(96,1.1483870355708736)(97,1.1610686775419667)(98,1.1865328799729358)(99,1.190629420410848)(100,1.1740331932163035)

};
\legend{Training Loss}
\end{axis}
\end{tikzpicture}
\hfill
\begin{tikzpicture}
\begin{axis}[
    width=1\textwidth,
    height=1\textwidth,
    xlabel={Epoch},
    ylabel={Loss},
    grid=major,
    legend pos=north east,
    ymin= .8,
    ymax= 2.8
]
\addplot[
    color=blue,
    thick
] coordinates {
(0,1.8581343963165053)(1,1.5586563730835332)(2,1.3740624054197017)(3,1.3622581070436253)(4,1.254717031444727)(5,1.2156354565930627)(6,1.2814604129442664)(7,1.458685939054586)(8,1.2502225729516816)(9,1.8546151319889643)(10,1.3932100969475594)(11,1.2237315778750257)(12,1.227602429179168)(13,1.1889620873443074)(14,1.1875061764133084)(15,1.2097010199140152)(16,1.1567439107620334)(17,1.2118164158073799)(18,1.1695099975489092)(19,1.1882815616923739)(20,1.1736812415008389)(21,1.1680192211749933)(22,1.1708597895050932)(23,1.1546354288931604)(24,1.15740282801213)(25,1.1249638809854827)(26,1.15115451381899)(27,1.1295865174747755)(28,1.2343661417115819)(29,1.3775926371511935)(30,1.2980754149923288)(31,1.2274241287360377)(32,1.3503148272253074)(33,1.3034850930611501)(34,1.275101758627492)(35,1.3332519019543267)(36,1.2868544597187703)(37,1.3498691799234457)(38,1.489728007594326)(39,1.3573322792591578)(40,1.2425792785220904)(41,1.2347844817830311)(42,1.3026488159453233)(43,1.4020728556066115)(44,1.3724089870665614)(45,1.4171333009170262)(46,1.236010891455867)(47,1.4520938471342275)(48,1.9618181813172488)(49,1.5977764756849433)(50,1.4655846722605725)(51,1.262578835517591)(52,1.3169490529304044)(53,1.3820507792697898)(54,1.3494177139174646)(55,1.3003703038262862)(56,1.3135480673605422)(57,1.4040331099076777)(58,1.2559256482616818)(59,1.3133652791909463)(60,1.440538624086072)(61,1.3578704646591824)(62,1.490120025228728)(63,1.6865046983352636)(64,1.6415722943968887)(65,1.3819694131999263)(66,1.4165461018789332)(67,1.3042649810367906)(68,1.3889723708978936)(69,1.5834751126129696)(70,1.3437626531518965)(71,1.2972069683604626)(72,1.3355535332992932)(73,1.26856460939043)(74,1.1932551980035153)(75,1.2299867320247204)(76,1.1855630208606078)(77,1.219163375268863)(78,1.2170681158236543)(79,1.1665665566687748)(80,1.200753418211767)(81,1.2067814079551753)(82,1.2238278057674465)(83,1.191209540221542)(84,1.214328292425917)(85,1.189616033295337)(86,1.187508109373293)(87,1.2664384576795258)(88,1.2958114928049111)(89,1.2714890256810059)(90,1.2479620653757921)(91,1.2105649960223472)(92,1.2450960571379666)(93,1.2057232101433046)(94,1.154757694859368)(95,1.1788566465097936)(96,1.1906744474057385)(97,1.153467999903242)(98,1.189184314382534)(99,1.209706147378588)(100,1.2336106774617983)

};
\legend{Training Loss}
\end{axis}
\end{tikzpicture}}
\caption{Graphical representation of Loss vs Epoch for training the TFT model for the Weather Prediction dataset: the left-hand side graph depicts the classical TFT model, the middle graph illustrates the quantum TFT model without a quantum LSTM,  and the right graph represents the quantum TFT model where the LSTM component is also quantum.}
\label{fig: loss vs epoch}
\end{figure*}

\subsection{Weather Prediction}
 In this section, we evaluate the performance of our proposed model by conducting experiments on a weather prediction task.
The dataset for this study is sourced from Kaggle and contains weather records covering from 2012-01-01 to 2015-12-31, i.e., a total of 1461 rows or instances.
Each instance is characterized by six features, namely: date, precipitation (all forms in which water falls on the land surface and open water bodies as rain, sleet, snow, hail, or drizzle), maximum temperature, minimum temperature, wind speed, and weather condition.
We utilize  26 instances collected from the year 2012-01-01, incorporating 4 input features: precipitation, maximum temperature, minimum temperature, and weather condition, with the wind speed as the target feature.


Table~\ref{Table 2} gives an overview of the classical Temporal Fusion Transformer and its quantum-based variants, one with no QLSTM and the other with a QLSTM. The table emphasizes important architectural parameters such as the LSTM hidden layer size, hidden dimension, quantum-specific units such as Angle Embedding for input encoding to provide an easy implementation, Basic Entangler Layers, a form of N-local circuits, but without the last rotation layer, for variational layers, and the measurement observable utilized Pauli-Z. It further compares the count of trainable parameters overall.

\begin{table*}[ht]
\centering
\resizebox{\textwidth}{!}{
\begin{tabular}{|l|c|c|c|}
\hline
\textbf{Parameter} & \textbf{Classical TFT} & \textbf{Quantum TFT (without QLSTM)} & \textbf{Quantum TFT (with QLSTM)} \\ 
\hline
LSTM Hidden Layer Size & 1 & 1 & QLSTM \\ 
\hline
Hidden Dimension & 4 & 4 & 4 \\ 
\hline
Input Encoding & -- & Angle Embedding & Angle Embedding \\ 
\hline
Variational Layer Type & -- & Basic Entangler Layers (N-local circuit) & Basic Entangler Layers (N-local circuit) \\ 
\hline
Variational Layer Depth (Layers) & -- & 4 & 4 \\ 
\hline
Measurement Observable & -- & Pauli-Z & Pauli-Z \\ 
\hline
Trainable Parameters & 282 & 236 & 252 \\ 
\hline
\end{tabular}}
\caption{Parameters for Classical and Quantum Temporal Fusion Transformer Configurations}
\label{Table 2}
\end{table*}

In Figure~\ref{fig: loss vs epoch}, we present the loss vs. epoch diagram for Classical TFT, QTFT without quantum LSTM, and QTFT with quantum LSTM for Weather Prediction. We take a total of 100 iteration steps for training and testing the model. Although both the QTFT model's graph exhibits more fluctuations compared to the classical TFT model, the overall result remains unaffected.

\begin{figure*}[htpb]
\begin{center}
\resizebox{.65\textwidth}{!}{\begin{tikzpicture}
\begin{axis}[
    width=\textwidth,
    height=\textwidth,
    title={Epoch 0},
    xlabel={Time},
    ylabel={Value},
    grid=major,
    legend pos=north east
]
\addplot[
    color=blue,
    thick
] coordinates {

(0,0.24210526049137115)(1,0.49473685026168823)(2,0.49473685026168823)(3,0.6421052813529968)(4,0.6421052813529968)(5,0.23157894611358643)(6,0.23157894611358643)(7,0.24210526049137115)(8,0.24210526049137115)(9,0.21052631735801697)(10,0.21052631735801697)(11,0.35789474844932556)(12,0.35789474844932556)(13,0.35789474844932556)(14,0.35789474844932556)(15,0.5368421077728271)(16,0.5368421077728271)(17,0.20000000298023224)(18,0.20000000298023224)(19,0.13684210181236267)(20,0.13684210181236267)(21,0.557894766330719)(22,0.557894766330719)(23,0.3368421196937561)(24,0.3368421196937561)(25,0.5263158082962036)(26,0.5263158082962036)(27,0.5894736647605896)(28,0.5894736647605896)(29,0.5263158082962036)(30,0.5263158082962036)(31,0.16842105984687805)(32,0.16842105984687805)(33,0.24210526049137115)

    };
    \addlegendentry{True Values}

\addplot[
    color=red,
    mark=x,
    mark size=5pt,
    thick
    ]
    coordinates {
    (0,0.37034785747528076)(1,0.4250416159629822)(2,0.42025595903396606)(3,0.2936651110649109)(4,0.2885212302207947)(5,0.26757150888442993)(6,0.2620317339897156)(7,0.28225743770599365)(8,0.2770146131515503)(9,0.29859432578086853)(10,0.29306018352508545)(11,0.3379995822906494)(12,0.3326210081577301)(13,0.25294357538223267)(14,0.24722586572170258)(15,0.3771398365497589)(16,0.37172257900238037)(17,0.3719784617424011)(18,0.3665841519832611)(19,0.36037570238113403)(20,0.3543740510940552)(21,0.5637710690498352)(22,0.5589065551757812)(23,0.539714515209198)(24,0.5350295901298523)(25,0.5361613035202026)(26,0.5312981009483337)(27,0.5634005069732666)(28,0.5589783787727356)(29,0.5811760425567627)(30,0.5768947601318359)(31,0.5621856451034546)(32,0.5578429698944092)(33,0.5847525000572205)

    };
    \addlegendentry{Predicted Values}
\end{axis}
\end{tikzpicture}

\vspace{0.05\textwidth}
\begin{tikzpicture}
\begin{axis}[
   width=\textwidth,
    height=\textwidth,
    title={Epoch 100},
    xlabel={Time},
    grid=major,
    legend pos=north east,
    yticklabels=\empty
]
\addplot[
    color=blue,
    thick
] coordinates {

(0,0.24210526049137115)(1,0.49473685026168823)(2,0.49473685026168823)(3,0.6421052813529968)(4,0.6421052813529968)(5,0.23157894611358643)(6,0.23157894611358643)(7,0.24210526049137115)(8,0.24210526049137115)(9,0.21052631735801697)(10,0.21052631735801697)(11,0.35789474844932556)(12,0.35789474844932556)(13,0.35789474844932556)(14,0.35789474844932556)(15,0.5368421077728271)(16,0.5368421077728271)(17,0.20000000298023224)(18,0.20000000298023224)(19,0.13684210181236267)(20,0.13684210181236267)(21,0.557894766330719)(22,0.557894766330719)(23,0.3368421196937561)(24,0.3368421196937561)(25,0.5263158082962036)(26,0.5263158082962036)(27,0.5894736647605896)(28,0.5894736647605896)(29,0.5263158082962036)(30,0.5263158082962036)(31,0.16842105984687805)(32,0.16842105984687805)(33,0.24210526049137115)

    };
    \addlegendentry{True Values}

\addplot[
    color=red,
    mark=x,
    mark size=5pt,
    thick
    ]
    coordinates {
    (0,0.3428930342197418)(1,0.38827523589134216)(2,0.3886059820652008)(3,0.2630399465560913)(4,0.2633829414844513)(5,0.22980263829231262)(6,0.230093851685524)(7,0.24733608961105347)(8,0.24760523438453674)(9,0.2676352262496948)(10,0.2679350674152374)(11,0.3029097616672516)(12,0.30319643020629883)(13,0.21869930624961853)(14,0.21907857060432434)(15,0.3335374593734741)(16,0.3336861729621887)(17,0.32987141609191895)(18,0.32991304993629456)(19,0.31777480244636536)(20,0.3184942901134491)(21,0.506600022315979)(22,0.5081650018692017)(23,0.4838199019432068)(24,0.48382067680358887)(25,0.48087963461875916)(26,0.4825595021247864)(27,0.5092152953147888)(28,0.5105944275856018)(29,0.5246256589889526)(30,0.5241559147834778)(31,0.5055757164955139)(32,0.5068672895431519)(33,0.5315248966217041)

    };
    \addlegendentry{Predicted Values}
\end{axis}
\end{tikzpicture}}
\noindent
\hspace{-0.5cm}
\resizebox{0.65\textwidth}{!}
{\begin{tikzpicture}
\begin{axis}[
    width=\textwidth,
    height=\textwidth,
    title={Epoch 0},
    xlabel={Time},
    ylabel={Value},
    grid=major,
    legend pos=north east
]
\addplot[
    color=blue,
    thick
] coordinates {

(0,0.24210526049137115)(1,0.49473685026168823)(2,0.49473685026168823)(3,0.6421052813529968)(4,0.6421052813529968)(5,0.23157894611358643)(6,0.23157894611358643)(7,0.24210526049137115)(8,0.24210526049137115)(9,0.21052631735801697)(10,0.21052631735801697)(11,0.35789474844932556)(12,0.35789474844932556)(13,0.35789474844932556)(14,0.35789474844932556)(15,0.5368421077728271)(16,0.5368421077728271)(17,0.20000000298023224)(18,0.20000000298023224)(19,0.13684210181236267)(20,0.13684210181236267)(21,0.557894766330719)(22,0.557894766330719)(23,0.3368421196937561)(24,0.3368421196937561)(25,0.5263158082962036)(26,0.5263158082962036)(27,0.5894736647605896)(28,0.5894736647605896)(29,0.5263158082962036)(30,0.5263158082962036)(31,0.16842105984687805)(32,0.16842105984687805)(33,0.24210526049137115)

    };
    \addlegendentry{True Values}

\addplot[
    color=red,
    mark=x,
    mark size=5pt,
    thick
    ]
    coordinates {
    (0,0.43369562614573376)(1,0.4646758454937125)(2,0.46033731749569584)(3,0.37989349894240476)(4,0.3780940914319517)(5,0.35948842653484925)(6,0.35765646320256894)(7,0.3730695865408905)(8,0.37129963957259815)(9,0.38800792962037467)(10,0.38382198708583176)(11,0.4120632053633577)(12,0.40890379949874534)(13,0.35376712854049785)(14,0.34890124664055916)(15,0.43716311744827063)(16,0.432411736474414)(17,0.4329388626270851)(18,0.4283962384949065)(19,0.42332893002945815)(20,0.41420460407257187)(21,0.5533331050388293)(22,0.5470184470598828)(23,0.5338009550580802)(24,0.5273980010235029)(25,0.5292054702463964)(26,0.522237266927492)(27,0.5431515847477076)(28,0.5397686020085082)(29,0.5505650210204819)(30,0.5473721421737554)(31,0.5392967940935776)(32,0.5357517196816073)(33,0.5550194921684705)
    };
    \addlegendentry{Predicted Values}
\end{axis}
\end{tikzpicture}

\vspace{0.05\textwidth}
\begin{tikzpicture}
\begin{axis}[
   width=\textwidth,
    height=\textwidth,
    title={Epoch 100},
    xlabel={Time},
    grid=major,
    legend pos=north east,
    yticklabels=\empty
]
\addplot[
    color=blue,
    thick
] coordinates {

(0,0.24210526049137115)(1,0.49473685026168823)(2,0.49473685026168823)(3,0.6421052813529968)(4,0.6421052813529968)(5,0.23157894611358643)(6,0.23157894611358643)(7,0.24210526049137115)(8,0.24210526049137115)(9,0.21052631735801697)(10,0.21052631735801697)(11,0.35789474844932556)(12,0.35789474844932556)(13,0.35789474844932556)(14,0.35789474844932556)(15,0.5368421077728271)(16,0.5368421077728271)(17,0.20000000298023224)(18,0.20000000298023224)(19,0.13684210181236267)(20,0.13684210181236267)(21,0.557894766330719)(22,0.557894766330719)(23,0.3368421196937561)(24,0.3368421196937561)(25,0.5263158082962036)(26,0.5263158082962036)(27,0.5894736647605896)(28,0.5894736647605896)(29,0.5263158082962036)(30,0.5263158082962036)(31,0.16842105984687805)(32,0.16842105984687805)(33,0.24210526049137115)
    };
    \addlegendentry{True Values}

\addplot[
    color=red,
    mark=x,
    mark size=5pt,
    thick
    ]
    coordinates {
    (0,0.3835133456125936)(1,0.43344499165125383)(2,0.43333619239888804)(3,0.32685147543595533)(4,0.3269992179319175)(5,0.3065209050897105)(6,0.3067908402979479)(7,0.3159221658250949)(8,0.3162941877104729)(9,0.32934416525029453)(10,0.3292555498467208)(11,0.35894875671387416)(12,0.35894548567128304)(13,0.2966441138799916)(14,0.29643898155353454)(15,0.3809705671535082)(16,0.3808334703497792)(17,0.3780631946490866)(18,0.3781295106501349)(19,0.3696181031291016)(20,0.3686221164238359)(21,0.5106366473361316)(22,0.5098785181823207)(23,0.4935853805491513)(24,0.4930748498676266)(25,0.48976986386781163)(26,0.48908591089563586)(27,0.5122827254086745)(28,0.5120667903364263)(29,0.5316307024043678)(30,0.531407374292841)(31,0.5151089184265509)(32,0.5148962503347615)(33,0.5296772162815875)

    };
    \addlegendentry{Predicted Values}
\end{axis}
\end{tikzpicture}}
\noindent
\hspace{-0.5cm}
\resizebox{0.65\textwidth}{!}{\begin{tikzpicture}
\begin{axis}[
    width=\textwidth,
    height=\textwidth,
    title={Epoch 0},
    xlabel={Time},
    ylabel={Value},
    grid=major,
    legend pos=north east
]
\addplot[
    color=blue,
    thick
] coordinates {
(0,0.24210526049137115)(1,0.49473685026168823)(2,0.49473685026168823)(3,0.6421052813529968)(4,0.6421052813529968)(5,0.23157894611358643)(6,0.23157894611358643)(7,0.24210526049137115)(8,0.24210526049137115)(9,0.21052631735801697)(10,0.21052631735801697)(11,0.35789474844932556)(12,0.35789474844932556)(13,0.35789474844932556)(14,0.35789474844932556)(15,0.5368421077728271)(16,0.5368421077728271)(17,0.20000000298023224)(18,0.20000000298023224)(19,0.13684210181236267)(20,0.13684210181236267)(21,0.557894766330719)(22,0.557894766330719)(23,0.3368421196937561)(24,0.3368421196937561)(25,0.5263158082962036)(26,0.5263158082962036)(27,0.5894736647605896)(28,0.5894736647605896)(29,0.5263158082962036)(30,0.5263158082962036)(31,0.16842105984687805)(32,0.16842105984687805)(33,0.24210526049137115)

    };
    \addlegendentry{True Values}

\addplot[
    color=red,
    mark=x,
    mark size=5pt,
    thick
    ]
    coordinates {
(0,0.1606238278621179)(1,0.1995267891540059)(2,0.19971907624730717)(3,0.08656607414380724)(4,0.08689647295250419)(5,0.05629043393365046)(6,0.05647743695649865)(7,0.07266284495036945)(8,0.07282760517926634)(9,0.09122181597131229)(10,0.0913210881454729)(11,0.12354712100630028)(12,0.12367247722867125)(13,0.04592669761191409)(14,0.04609427683584477)(15,0.14678556347114025)(16,0.14684032962670812)(17,0.14327402227145436)(18,0.14349229904105149)(19,0.13218374956622875)(20,0.13225188600120968)(21,0.3014028470696101)(22,0.30149176719509696)(23,0.2798305980241739)(24,0.28010430356346117)(25,0.2776144391793941)(26,0.27781664463387057)(27,0.3020887798861991)(28,0.3025064912819322)(29,0.3150744234901838)(30,0.31549309132765396)(31,0.2988789476013489)(32,0.2992799160589074)(33,0.32180951039902267)

    };
    \addlegendentry{Predicted Values}
\end{axis}
\end{tikzpicture}

\vspace{0.05\textwidth}
\begin{tikzpicture}
\begin{axis}[
   width=\textwidth,
    height=\textwidth,
    title={Epoch 100},
    xlabel={Time},
    grid=major,
    legend pos=north east,
    yticklabels=\empty
]
\addplot[
    color=blue,
    thick
] coordinates {
     (0,0.24210526049137115)(1,0.49473685026168823)(2,0.49473685026168823)(3,0.6421052813529968)(4,0.6421052813529968)(5,0.23157894611358643)(6,0.23157894611358643)(7,0.24210526049137115)(8,0.24210526049137115)(9,0.21052631735801697)(10,0.21052631735801697)(11,0.35789474844932556)(12,0.35789474844932556)(13,0.35789474844932556)(14,0.35789474844932556)(15,0.5368421077728271)(16,0.5368421077728271)(17,0.20000000298023224)(18,0.20000000298023224)(19,0.13684210181236267)(20,0.13684210181236267)(21,0.557894766330719)(22,0.557894766330719)(23,0.3368421196937561)(24,0.3368421196937561)(25,0.5263158082962036)(26,0.5263158082962036)(27,0.5894736647605896)(28,0.5894736647605896)(29,0.5263158082962036)(30,0.5263158082962036)(31,0.16842105984687805)(32,0.16842105984687805)(33,0.24210526049137115)
    };
    \addlegendentry{True Values}

\addplot[
    color=red,
    mark=x,
    mark size=5pt,
    thick
    ]
    coordinates {
(0,0.4471546497835922)(1,0.47573046894953297)(2,0.4723066928343829)(3,0.36038157563167816)(4,0.35649020979842244)(5,0.33045551625756864)(6,0.32728884985903467)(7,0.34678663739083165)(8,0.34371698345245666)(9,0.3625277751758718)(10,0.35974432300499326)(11,0.4036379726035511)(12,0.4000475672632048)(13,0.3137562416693056)(14,0.31084975015896593)(15,0.4048084074690851)(16,0.40202526331098215)(17,0.4006748769392954)(18,0.3970628729654743)(19,0.3876579740878644)(20,0.38566022952959356)(21,0.5630865839322943)(22,0.5601787250343309)(23,0.5356432813912789)(24,0.5321396091513797)(25,0.5340955507241222)(26,0.5307884326156966)(27,0.5642831593779316)(28,0.5609993715391531)(29,0.5723417099718588)(30,0.5689563352628466)(31,0.5563720775991825)(32,0.5530783436204105)(33,0.5816359385001766)

    };
    \addlegendentry{Predicted Values}
\end{axis}
\end{tikzpicture}}
\caption{Learning from the Weather Prediction dataset: the top row depicts the close values vs. the time series graph for the classical TFT model, 
the middle row represents the quantum TFT without quantum LSTM, 
and the bottom row corresponds to the quantum TFT model with quantum LSTM components. At Epoch 0,  the model calculates the loss and applies an optimizer for backpropagation as an initial step. Similarly, Epoch 100 indicates that the model computed the loss and applied the optimizer across 100 steps from the initial step.}
\label{fig: Time vs value}
\end{center}
\end{figure*}
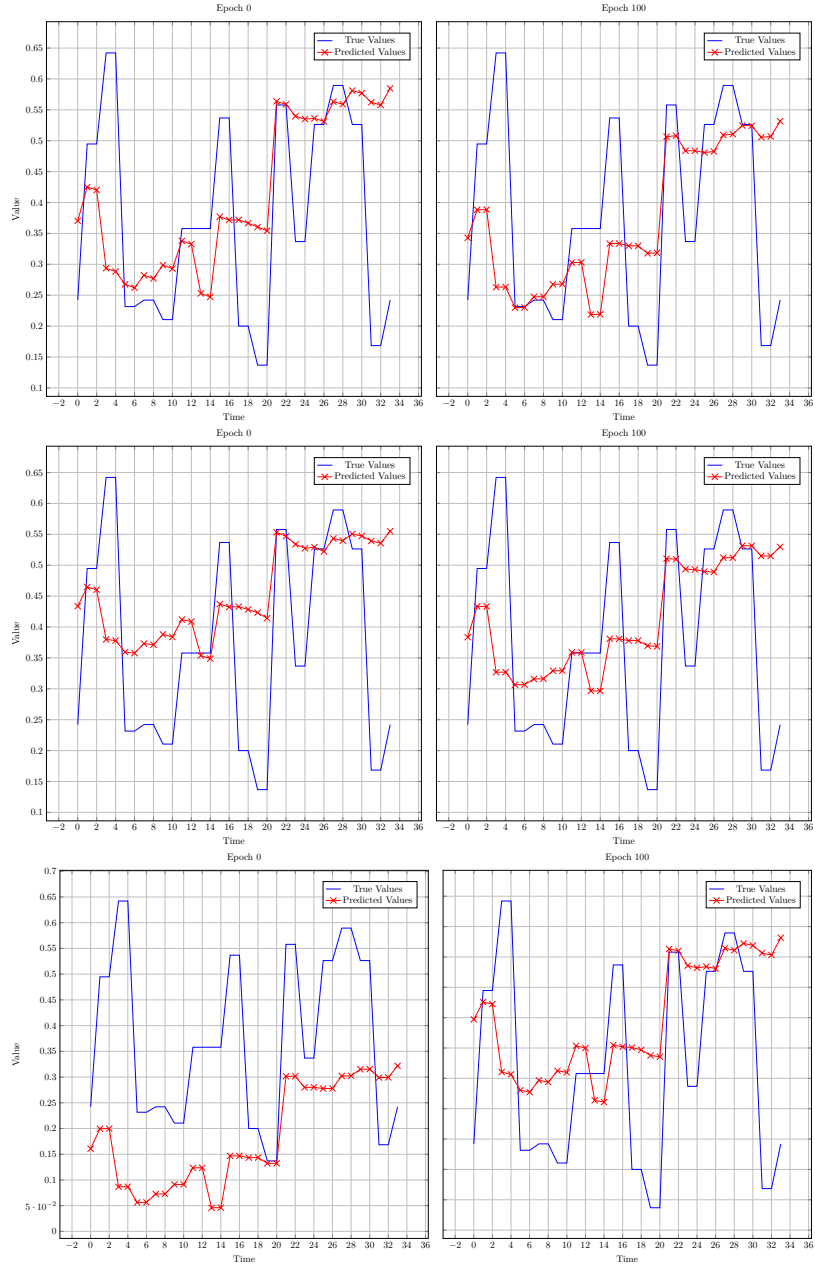

Figure~\ref{fig: Time vs value} illustrates the training behaviors on the Weather Prediction dataset of both the quantum and the classical TFT models during epoch 0 to epoch 100.
In the diagram, the blue line represents the true closing value over time, while the dotted red line corresponds to the predicted closing values. 
An inspection reveals that the graph shows almost the same close values for two consecutive time steps, except at the start and end.
This setup defines the configuration of our model, where we use a window of two past time steps to predict two future time steps.  
As a result of this overlapping window approach, we present almost the same predicted close values for two consecutive time steps, except for the starting and ending points.

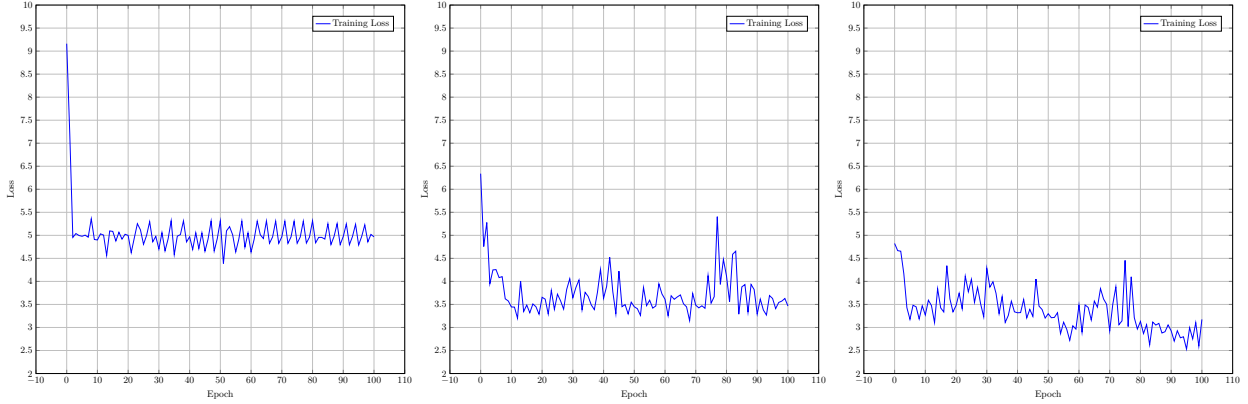
\begin{figure*}[htpb]
\begin{center}
\resizebox{\textwidth}{!}{
\begin{tikzpicture}
\begin{axis}[
    width=1\textwidth,
    height=1\textwidth,
    xlabel={Epoch},
    ylabel={Loss},
    grid=major,
    legend pos=north east,
    ymin=2,
    ymax=10
]
\addplot[
    color=blue,
    thick
] coordinates {
    (0,9.161436080932617)(1,7.207069396972656)(2,4.956055164337158)(3,5.038845539093018)(4,4.996583461761475)(5,4.980082988739014)(6,5.004107475280762)(7,4.960165977478027)(8,5.359270095825195)(9,4.910433292388916)(10,4.899941921234131)(11,5.032714366912842)(12,5.005336284637451)(13,4.564659595489502)(14,5.096315383911133)(15,5.085684299468994)(16,4.877832889556885)(17,5.066482067108154)(18,4.917731285095215)(19,5.023671627044678)(20,4.998227119445801)(21,4.625833988189697)(22,4.94356632232666)(23,5.252756118774414)(24,5.119686603546143)(25,4.806312084197998)(26,4.999327659606934)(27,5.296051979064941)(28,4.860052108764648)(29,4.976824760437012)(30,4.702857971191406)(31,5.05821418762207)(32,4.66446590423584)(33,4.922759532928467)(34,5.316331386566162)(35,4.582251071929932)(36,4.98257303237915)(37,5.01602840423584)(38,5.313950061798096)(39,4.857069492340088)(40,4.968907833099365)(41,4.695046424865723)(42,5.056216239929199)(43,4.714295387268066)(44,5.053745746612549)(45,4.656649589538574)(46,4.917806148529053)(47,5.31164026260376)(48,4.6590189933776855)(49,4.929831027984619)(50,5.315188407897949)(51,4.38918399810791)(52,5.097573757171631)(53,5.188589572906494)(54,5.013421535491943)(55,4.638981342315674)(56,4.91408109664917)(57,5.307642459869385)(58,4.743530750274658)(59,5.053263187408447)(60,4.638916969299316)(61,4.913498878479004)(62,5.3066725730896)(63,5.011746406555176)(64,4.931375503540039)(65,5.313374042510986)(66,4.8272199630737305)(67,4.965882778167725)(68,5.308056354522705)(69,4.826287269592285)(70,4.967221260070801)(71,5.304402828216553)(72,4.826452732086182)(73,4.9665446281433105)(74,5.3044514656066895)(75,4.828214168548584)(76,4.964354515075684)(77,5.30800724029541)(78,4.831826210021973)(79,4.9607319831848145)(80,5.3155317306518555)(81,4.837868690490723)(82,4.955470561981201)(83,4.9554762840271)(84,4.918903350830078)(85,5.249971866607666)(86,4.801409721374512)(87,4.975937843322754)(88,5.253529071807861)(89,4.799881458282471)(90,4.975254535675049)(91,5.248401165008545)(92,4.795770645141602)(93,4.975606918334961)(94,5.240203380584717)(95,4.7899017333984375)(96,4.977352142333984)(97,5.22825288772583)(98,4.858888149261475)(99,5.01967191696167)(100,4.9680023193359375)

};
 
\legend{Training Loss}
\end{axis}
\end{tikzpicture}
\hfill
\begin{tikzpicture}
\begin{axis}[
    width=1\textwidth,
    height=1\textwidth,
    xlabel={Epoch},
    ylabel={Loss},
    grid=major,
    legend pos=north east,
    ymin=2,
    ymax=10
]
\addplot[
    color=blue,
    thick
] coordinates {

(0,6.339681417218581)(1,4.7574477987247565)(2,5.27912002930473)(3,3.9508253089133314)(4,4.250162054953767)(5,4.256377320543911)(6,4.0830168586986515)(7,4.1031912569212246)(8,3.6226047400042223)(9,3.57473814712837)(10,3.4455491144341828)(11,3.44222400567622)(12,3.2011251999220063)(13,4.007612752812316)(14,3.3475236149469643)(15,3.486712193193111)(16,3.3194248839598393)(17,3.5107065499031878)(18,3.450215857682193)(19,3.287830091707481)(20,3.655348568755148)(21,3.6150106720092827)(22,3.291967780278834)(23,3.8036289903508598)(24,3.394312580673522)(25,3.723297126840034)(26,3.579428199100814)(27,3.39815424611964)(28,3.8270854514910013)(29,4.057557349215693)(30,3.6464198996093717)(31,3.8686081146926297)(32,4.0313324675821764)(33,3.384259286446759)(34,3.7622936121500876)(35,3.67341844054479)(36,3.489082652383768)(37,3.3876666069223482)(38,3.759622514862115)(39,4.2622608723297155)(40,3.6344645767205215)(41,3.8914227417735257)(42,4.523313561581928)(43,3.7814248739926812)(44,3.2991203348169345)(45,4.222746313391049)(46,3.451474233930612)(47,3.4965314036849033)(48,3.294669762903702)(49,3.5494172783290576)(50,3.449017361960233)(51,3.4077330919684776)(52,3.262496041224618)(53,3.869082508753878)(54,3.4736683842926404)(55,3.591966131051623)(56,3.4200006802571608)(57,3.4674072730994094)(58,3.9474097610120626)(59,3.7276384292665075)(60,3.611924852031578)(61,3.2556798732884102)(62,3.6883979149430344)(63,3.611357334338826)(64,3.663287284031669)(65,3.706897212543246)(66,3.5151424656585872)(67,3.447604265136797)(68,3.1486706499800166)(69,3.7373049481793936)(70,3.4832382492017624)(71,3.425913029362402)(72,3.470478113520427)(73,3.4170927453125985)(74,4.124979320456772)(75,3.5277192339397354)(76,3.6691546653465137)(77,5.405647623089145)(78,3.9363142216224)(79,4.476634280085184)(80,4.119515928113302)(81,3.5587939548711187)(82,4.5898629064963705)(83,4.656174925455381)(84,3.288602742204306)(85,3.881534158518474)(86,3.9313932535969194)(87,3.333611442016001)(88,3.933740206340734)(89,3.8237947738108584)(90,3.291459159776183)(91,3.615528271571945)(92,3.3788444323016957)(93,3.2695720459699515)(94,3.6929399138360717)(95,3.629305804413778)(96,3.409697227389051)(97,3.5434992435107135)(98,3.572672368319945)(99,3.6290974982228486)(100,3.464645147792295)

};
\legend{Training Loss}
\end{axis}
\end{tikzpicture}
\hfill
\begin{tikzpicture}
\begin{axis}[
    width=1\textwidth,
    height=1\textwidth,
    xlabel={Epoch},
    ylabel={Loss},
    grid=major,
    legend pos=north east,
    ymin=2,
    ymax=10
]
\addplot[
    color=blue,
    thick
] coordinates {
(0,4.823968256263786)
(1,4.668553139210827)
(2,4.65573188314054)
(3,4.168642315480399)
(4,3.4298060148460783)
(5,3.1764936518673252)
(6,3.4822407019074566)
(7,3.4489967133176913)
(8,3.1844787812151862)
(9,3.470002043215467)
(10,3.2705743905481173)
(11,3.591693314795985)
(12,3.474759654567082)
(13,3.105967809156457)
(14,3.841306904359496)
(15,3.4253286270931254)
(16,3.3363306468848934)
(17,4.341538780785249)
(18,3.619694942850632)
(19,3.339603652856428)
(20,3.4747805655367427)
(21,3.7416746552536777)
(22,3.407818839364494)
(23,4.113489196150317)
(24,3.7757843858258804)
(25,4.046998180178333)
(26,3.545777425153437)
(27,3.8711494909807156)
(28,3.51213604330053)
(29,3.230552733447966)
(30,4.283619338380664)
(31,3.8754797562933936)
(32,3.9919927699239164)
(33,3.7431339050763146)
(34,3.3007188072820286)
(35,3.6704013727953626)
(36,3.110367356381812)
(37,3.2603701444262603)
(38,3.5707478773401506)
(39,3.339856635436882)
(40,3.319189578076105)
(41,3.3234130739558374)
(42,3.606848613298407)
(43,3.2067891511312085)
(44,3.3920327498417144)
(45,3.2411437206981253)
(46,4.05165327026901)
(47,3.463251198017126)
(48,3.3895243304436296)
(49,3.2043186818555833)
(50,3.2983812800361596)
(51,3.211860459550987)
(52,3.2185358619880784)
(53,3.3200910646266655)
(54,2.866220839379812)
(55,3.1144768078987326)
(56,2.964085191050013)
(57,2.721287242569306)
(58,3.036504079846292)
(59,2.9647440450312956)
(60,3.4859060495575136)
(61,2.8976395470198577)
(62,3.487701631267412)
(63,3.434914174458081)
(64,3.1618684002040425)
(65,3.5722252372008985)
(66,3.4403801166617365)
(67,3.8437636031967948)
(68,3.618129955089919)
(69,3.503065536687605)
(70,2.9271304580472757)
(71,3.5033793239164286)
(72,3.881412534130548)
(73,3.060899304904515)
(74,3.144004681718429)
(75,4.457210184976724)
(76,3.0200534976257964)
(77,4.10005888203831)
(78,3.213567362325163)
(79,2.9661592162667114)
(80,3.1307121507662288)
(81,2.8681738216245805)
(82,3.059759561789612)
(83,2.6233738883681648)
(84,3.1169725243449875)
(85,3.05250626961326)
(86,3.088470778830338)
(87,2.87984630960264)
(88,2.908153104502164)
(89,3.055172496459888)
(90,2.927865599960981)
(91,2.7036716072323292)
(92,2.9282420437999885)
(93,2.776575472141295)
(94,2.797770936347624)
(95,2.530399578316768)
(96,2.99572729572305)
(97,2.7518733235923527)
(98,3.1046965600139114)
(99,2.5949334832083366)
(100,3.175095362820266)
};
\legend{Training Loss}
\end{axis}
\end{tikzpicture}}
\caption{Graphical representation of Loss vs Epoch for training the TFT model for the Axis Bank dataset: the left-hand side graph depicts the classical TFT model, the middle graph illustrates the quantum TFT model without a quantum LSTM,  and the right graph represents the quantum TFT model where the LSTM component is also quantum.}
\label{fig: loss vs epoch2}
\end{center}
\end{figure*}

Table~\ref{Table 3} presents the training and testing loss values for Weather Prediction,  computed using the quantile loss function, as described earlier. Each value represents the average loss over a sliding window, using the same setup described above. From this setup, it can be observed that the QTFT model with quantum LSTM achieves the lowest loss, outperforming both the  QTFT model without quantum LSTM and, lastly, the classical TFT model.

\begin{table*}[htpb]
\centering
\resizebox{0.6\textwidth}{!}{
\begin{tabular}{|l|c|c|}
\hline
    Model & Training Loss & Testing Loss \\ \hline
TFT  & \(0.2622 \) & \(0.1066\) \\ \hline
QTFT (Without QLSTM)  & \(0.0575\) & \(0.0997 \) \\ \hline
QTFT (With QLSTM)  & \(0.0594 \) & \(0.0868\) \\ \hline
\end{tabular}}
\caption{Comparison of training and testing loss values for Weather Prediction dataset.}
\label{Table 3}
\end{table*}

\subsection{Stock Market Prediction}

We investigate the capability of our QTFT in learning the stock market and efficiently predicting. In this section, we pick the stock market data of Axis Bank from Kaggle and analyze it using the QTFT model.    
It represents a Nifty-50 stock market data record covering the years 2000 to 2021. The data set comes with 5306 rows or instances and 15 columns or features, describing various aspects of the stock data, including date, symbol, series, previous close, open, high, low, last, close, vwap, turnover, trades, deliverable volume, and deliverable percent.
For limitations of quantum hardware, we select the first 26 instances starting from the year 2000 and focus on 4 input features: open, high, low, and last, with the close price as the target feature.
The intention for using these selected feature vectors and features lies in their typically small numerical values, which facilitate more efficient QTFT model training and testing.

Table~\ref{Table 4} presents a comparison between the classical Temporal Fusion Transformer and its quantum-enhanced counterparts, one without a quantum LSTM (QLSTM) and another incorporating a QLSTM, which follows a similar experimental setup to that used for weather prediction.

In Figure~\ref{fig: loss vs epoch2}, we present the loss versus epoch curve for the classical TFT, QTFT without quantum LSTM, and QTFT with quantum LSTM on the Axis Bank dataset, which follows a similar kind of diagram as Weather Prediction.

\begin{table*}[htpb]
\centering
\resizebox{\textwidth}{!}{
\begin{tabular}{|l|c|c|c|}
\hline
\textbf{Parameter} & \textbf{Classical TFT} & \textbf{Quantum TFT (without QLSTM)} & \textbf{Quantum TFT (with QLSTM)} \\ 
\hline
LSTM Hidden Layer Size & 1 & 1 & QLSTM \\ 
\hline
Hidden Dimension & 2 & 2 & 2 \\ 
\hline
Input Encoding & -- & Angle Embedding & Angle Embedding \\ 
\hline
Variational Layer Type & -- & Basic Entangler Layers (N-local circuit) & Basic Entangler Layers (N-local circuit) \\ 
\hline
Variational Layer Depth (Layers) & -- & 2 & 2 \\ 
\hline
Measurement Observable & -- & Pauli-Z & Pauli-Z \\ 
\hline
Trainable Parameters & 190 & 158 & 174 \\ 
\hline
\end{tabular}}
\caption{Parameters for Classical and Quantum Temporal Fusion Transformer Configurations}
\label{Table 4}
\end{table*}

\begin{figure*}[htpb]
\begin{center}

\resizebox{0.65\textwidth}{!}{\begin{tikzpicture}
\begin{axis}[
    width=\textwidth,
    height=\textwidth,
    title={Epoch 0},
    xlabel={Time},
    ylabel={Value},
    grid=major,
    legend pos=north east
]
\addplot[
    color=blue,
    thick
] coordinates {

(0,26.299999237060547)(1,25.950000762939453)(2,25.950000762939453)(3,24.799999237060547)(4,24.799999237060547)(5,25.0)(6,25.0)(7,23.200000762939453)(8,23.200000762939453)(9,24.0)(10,24.0)(11,23.600000381469727)(12,23.600000381469727)(13,23.25)(14,23.25)(15,25.149999618530273)(16,25.149999618530273)(17,24.899999618530273)(18,24.899999618530273)(19,25.600000381469727)(20,25.600000381469727)(21,24.450000762939453)(22,24.450000762939453)(23,25.100000381469727)(24,25.100000381469727)(25,24.799999237060547)(26,24.799999237060547)(27,25.049999237060547)(28,25.049999237060547)(29,27.049999237060547)(30,27.049999237060547)(31,29.25)(32,29.25)(33,31.600000381469727)
    };
    \addlegendentry{True Values}

\addplot[
    color=red,
    mark=x,
    mark size=5pt,
    thick
    ]
    coordinates {
    (0,27.875537872314453)(1,27.127723693847656)(2,27.12690544128418)(3,26.129268646240234)(4,26.128448486328125)(5,26.628538131713867)(6,26.62771987915039)(7,24.93062400817871)(8,24.9298038482666)(9,24.63086700439453)(10,24.630048751831055)(11,24.63086700439453)(12,24.630048751831055)(13,24.33106803894043)(14,24.330249786376953)(15,25.280290603637695)(16,25.279470443725586)(17,26.628538131713867)(18,26.62771987915039)(19,26.129268646240234)(20,26.128448486328125)(21,26.37891387939453)(22,26.378095626831055)(23,25.430130004882812)(24,25.429311752319336)(25,25.87959861755371)(26,25.878780364990234)(27,25.530019760131836)(28,25.52920150756836)(29,27.177637100219727)(30,27.17681884765625)(31,29.37317657470703)(32,29.372358322143555)(33,31.71732521057129)

    };
    \addlegendentry{Predicted Values}
\end{axis}
\end{tikzpicture}

\vspace{0.05\textwidth}
\begin{tikzpicture}
\begin{axis}[
   width=\textwidth,
    height=\textwidth,
    title={Epoch 100},
    xlabel={Time},
    grid=major,
    legend pos=north east,
    yticklabels=\empty
]
\addplot[
    color=blue,
    thick
] coordinates {

(0,26.299999237060547)(1,25.950000762939453)(2,25.950000762939453)(3,24.799999237060547)(4,24.799999237060547)(5,25.0)(6,25.0)(7,23.200000762939453)(8,23.200000762939453)(9,24.0)(10,24.0)(11,23.600000381469727)(12,23.600000381469727)(13,23.25)(14,23.25)(15,25.149999618530273)(16,25.149999618530273)(17,24.899999618530273)(18,24.899999618530273)(19,25.600000381469727)(20,25.600000381469727)(21,24.450000762939453)(22,24.450000762939453)(23,25.100000381469727)(24,25.100000381469727)(25,24.799999237060547)(26,24.799999237060547)(27,25.049999237060547)(28,25.049999237060547)(29,27.049999237060547)(30,27.049999237060547)(31,29.25)(32,29.25)(33,31.600000381469727)
    };
    \addlegendentry{True Values}

\addplot[
    color=red,
    mark=x,
    mark size=5pt,
    thick
    ]
    coordinates {
    (0,26.946760177612305)(1,26.197952270507812)(2,26.196760177612305)(3,25.197952270507812)(4,25.196760177612305)(5,25.697952270507812)(6,25.696760177612305)(7,23.99795150756836)(8,23.99675941467285)(9,23.697952270507812)(10,23.696760177612305)(11,23.697952270507812)(12,23.696760177612305)(13,23.397953033447266)(14,23.396760940551758)(15,24.347951889038086)(16,24.346759796142578)(17,25.697952270507812)(18,25.696760177612305)(19,25.197952270507812)(20,25.196760177612305)(21,25.447952270507812)(22,25.446760177612305)(23,24.49795150756836)(24,24.49675941467285)(25,24.947952270507812)(26,24.946760177612305)(27,24.597951889038086)(28,24.596759796142578)(29,26.24795150756836)(30,26.24675941467285)(31,28.447952270507812)(32,28.446760177612305)(33,30.79795265197754)

    };
    \addlegendentry{Predicted Values}
\end{axis}
\end{tikzpicture}}
\noindent
\hspace{-0.5cm}
\resizebox{0.65\textwidth}{!}
{\begin{tikzpicture}
\begin{axis}[
    width=\textwidth,
    height=\textwidth,
    title={Epoch 0},
    xlabel={Time},
    ylabel={Value},
    grid=major,
    legend pos=north east
]
\addplot[
    color=blue,
    thick
] coordinates {

(0,26.299999237060547)(1,25.950000762939453)(2,25.950000762939453)(3,24.799999237060547)(4,24.799999237060547)(5,25.0)(6,25.0)(7,23.200000762939453)(8,23.200000762939453)(9,24.0)(10,24.0)(11,23.600000381469727)(12,23.600000381469727)(13,23.25)(14,23.25)(15,25.149999618530273)(16,25.149999618530273)(17,24.899999618530273)(18,24.899999618530273)(19,25.600000381469727)(20,25.600000381469727)(21,24.450000762939453)(22,24.450000762939453)(23,25.100000381469727)(24,25.100000381469727)(25,24.799999237060547)(26,24.799999237060547)(27,25.049999237060547)(28,25.049999237060547)(29,27.049999237060547)(30,27.049999237060547)(31,29.25)(32,29.25)(33,31.600000381469727)
    };
    \addlegendentry{True Values}

\addplot[
    color=red,
    mark=x,
    mark size=5pt,
    thick
    ]
    coordinates {
    (0,26.9588772010424)(1,26.372183756235504)(2,26.367116718335904)(3,25.579347284848705)(4,25.57947907746286)(5,25.751316158235355)(6,25.746363151227953)(7,24.469172701776362)(8,24.524519669546898)(9,23.970439980132856)(10,24.020008006296184)(11,24.16123244480365)(12,24.195427806957934)(13,23.807024727005473)(14,23.800100223606677)(15,24.984996389827675)(16,24.985842622402384)(17,25.956554335146627)(18,25.965540921426133)(19,25.506427350638987)(20,25.51011863735586)(21,25.74823908808153)(22,25.748653467112575)(23,25.063316917392644)(24,25.07338746769037)(25,25.29989498822185)(26,25.300137355493277)(27,25.013105420713)(28,24.99560048713269)(29,26.735747041855795)(30,26.733983516142626)(31,28.574108869168526)(32,28.5895117106612)(33,31.41884907582373)

    };
    \addlegendentry{Predicted Values}
\end{axis}
\end{tikzpicture}

\vspace{0.05\textwidth}
\begin{tikzpicture}
\begin{axis}[
   width=\textwidth,
    height=\textwidth,
    title={Epoch 100},
    xlabel={Time},
    grid=major,
    legend pos=north east,
    yticklabels=\empty
]
\addplot[
    color=blue,
    thick
] coordinates {

     (0,26.299999237060547)(1,25.950000762939453)(2,25.950000762939453)(3,24.799999237060547)(4,24.799999237060547)(5,25.0)(6,25.0)(7,23.200000762939453)(8,23.200000762939453)(9,24.0)(10,24.0)(11,23.600000381469727)(12,23.600000381469727)(13,23.25)(14,23.25)(15,25.149999618530273)(16,25.149999618530273)(17,24.899999618530273)(18,24.899999618530273)(19,25.600000381469727)(20,25.600000381469727)(21,24.450000762939453)(22,24.450000762939453)(23,25.100000381469727)(24,25.100000381469727)(25,24.799999237060547)(26,24.799999237060547)(27,25.049999237060547)(28,25.049999237060547)(29,27.049999237060547)(30,27.049999237060547)(31,29.25)(32,29.25)(33,31.600000381469727)
    };
    \addlegendentry{True Values}

\addplot[
    color=red,
    mark=x,
    mark size=5pt,
    thick
    ]
    coordinates {
    (0,26.77878630224207)(1,25.728268709330408)(2,25.733938789782435)(3,25.068215092557246)(4,25.065980941140566)(5,25.157102337645775)(6,25.1832313135447)(7,24.38226201987542)(8,24.38968991563064)(9,24.035696268694075)(10,24.018280045013434)(11,24.070390782502603)(12,24.043353801206752)(13,23.770218254154866)(14,23.773616696736102)(15,24.70184618108946)(16,24.737401400297106)(17,25.26745471259266)(18,25.295019369629795)(19,25.07898325686044)(20,25.084846715773104)(21,25.180793295847355)(22,25.18291673678687)(23,24.833493337723276)(24,24.837284390551304)(25,24.927657624313024)(26,24.92043719273113)(27,24.763404645108604)(28,24.775800354754704)(29,26.416978528730933)(30,26.48532139140631)(31,29.396313868364537)(32,29.380391283948267)(33,31.111919686761528)

    };
    \addlegendentry{Predicted Values}
\end{axis}
\end{tikzpicture}}
\noindent
\hspace{-0.5cm}
\resizebox{0.65\textwidth}{!}{\begin{tikzpicture}
\begin{axis}[
    width=\textwidth,
    height=\textwidth,
    title={Epoch 0},
    xlabel={Time},
    ylabel={Value},
    grid=major,
    legend pos=north east
]
\addplot[
    color=blue,
    thick
] coordinates {
(0,26.299999237060547)
(1,25.950000762939453)
(2,25.950000762939453)
(3,24.799999237060547)
(4,24.799999237060547)
(5,25.0)
(6,25.0)
(7,23.200000762939453)
(8,23.200000762939453)
(9,24.0)
(10,24.0)
(11,23.600000381469727)
(12,23.600000381469727)
(13,23.25)
(14,23.25)
(15,25.149999618530273)
(16,25.149999618530273)
(17,24.899999618530273)
(18,24.899999618530273)
(19,25.600000381469727)
(20,25.600000381469727)
(21,24.450000762939453)
(22,24.450000762939453)
(23,25.100000381469727)
(24,25.100000381469727)
(25,24.799999237060547)
(26,24.799999237060547)
(27,25.049999237060547)
(28,25.049999237060547)
(29,27.049999237060547)
(30,27.049999237060547)
(31,29.25)
(32,29.25)
(33,31.600000381469727)

    };
    \addlegendentry{True Values}

\addplot[
    color=red,
    mark=x,
    mark size=5pt,
    thick
    ]
    coordinates {
(0,26.146551572093077)
(1,25.45769166755922)
(2,25.45328980855236)
(3,24.85881123320916)
(4,24.904791073593522)
(5,25.015602558398076)
(6,25.044926255797034)
(7,24.673926801097693)
(8,24.662745997546363)
(9,24.31048946883879)
(10,24.291625787559223)
(11,24.44584088545159)
(12,24.331035114435355)
(13,24.10038085689226)
(14,24.096020293060796)
(15,24.537686073657074)
(16,24.76214926509246)
(17,25.21570590173293)
(18,25.2236198621099)
(19,24.928937855716512)
(20,24.9628614005632)
(21,25.071597377908528)
(22,25.07184530146868)
(23,24.887104001346543)
(24,24.881068096896925)
(25,24.949083825169634)
(26,24.925888179958)
(27,24.86869262590093)
(28,24.873553010706942)
(29,25.81211938325047)
(30,25.79945373195905)
(31,29.05099744530768)
(32,29.05939815228032)
(33,30.95791277011986)

    };
    \addlegendentry{Predicted Values}
\end{axis}
\end{tikzpicture}

\vspace{0.05\textwidth}
\begin{tikzpicture}
\begin{axis}[
   width=\textwidth,
    height=\textwidth,
    title={Epoch 100},
    xlabel={Time},
    grid=major,
    legend pos=north east,
    yticklabels=\empty
]
\addplot[
    color=blue,
    thick
] coordinates {
     (0,26.299999237060547)(1,25.950000762939453)(2,25.950000762939453)(3,24.799999237060547)(4,24.799999237060547)(5,25.0)(6,25.0)(7,23.200000762939453)(8,23.200000762939453)(9,24.0)(10,24.0)(11,23.600000381469727)(12,23.600000381469727)(13,23.25)(14,23.25)(15,25.149999618530273)(16,25.149999618530273)(17,24.899999618530273)(18,24.899999618530273)(19,25.600000381469727)(20,25.600000381469727)(21,24.450000762939453)(22,24.450000762939453)(23,25.100000381469727)(24,25.100000381469727)(25,24.799999237060547)(26,24.799999237060547)(27,25.049999237060547)(28,25.049999237060547)(29,27.049999237060547)(30,27.049999237060547)(31,29.25)(32,29.25)(33,31.600000381469727)
    };
    \addlegendentry{True Values}

\addplot[
    color=red,
    mark=x,
    mark size=5pt,
    thick
    ]
    coordinates {
(0,26.37512363920779)
(1,25.8517837463664)
(2,25.851553908089798)
(3,25.093265116324176)
(4,25.09111667138122)
(5,25.21539885546761)
(6,25.204189772430873)
(7,24.126211951041334)
(8,24.115153078642276)
(9,23.90187043529048)
(10,23.89210349406266)
(11,23.909629505097673)
(12,23.899982296316598)
(13,23.571581138420207)
(14,23.57557397345916)
(15,24.814654915436165)
(16,24.829637874802497)
(17,25.388080686328653)
(18,25.391038842348344)
(19,25.186347767749794)
(20,25.18788418937473)
(21,25.196766358316243)
(22,25.194612665833766)
(23,24.860837612770585)
(24,24.85727800530493)
(25,24.91121791484768)
(26,24.908896837242924)
(27,24.75888573019655)
(28,24.75982833625842)
(29,26.582613504078417)
(30,26.59068987734903)
(31,29.45594144556574)
(32,29.449715977465054)
(33,31.228180986485334)

    };
    \addlegendentry{Predicted Values}
\end{axis}
\end{tikzpicture}}
\caption{Learning from the Axis Bank dataset: the top row depicts the close values vs. the time series graph for the classical TFT model, 
the middle row represents the quantum TFT without quantum LSTM, 
and the bottom row corresponds to the quantum TFT model with quantum LSTM components. At Epoch 0,  the model calculates the loss and applies an optimizer for backpropagation as an initial step. Similarly, Epoch 100 indicates that the model computed the loss and applied the optimizer across 100 steps from the initial step.}
\label{fig: Time vs value 2}
\end{center}
\end{figure*}
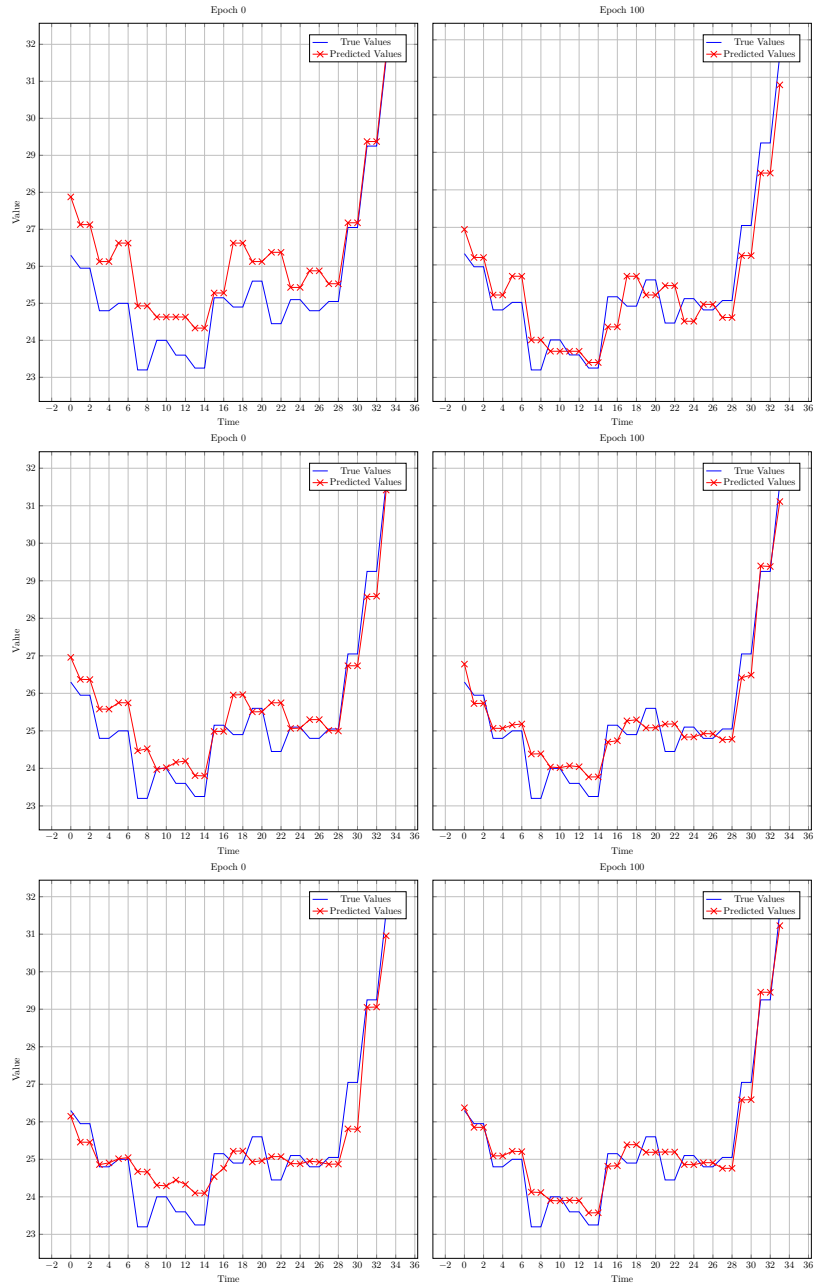

Figure~\ref{fig: Time vs value 2} illustrates the training behaviors on the Axis bank dataset of both the quantum and the classical TFT models during epoch 0 to epoch 100, which follows a similar kind of diagram as weather prediction.

Table~\ref{Table 5} presents the training and testing loss values for the Axis Bank dataset, which follows a similar experimental result as Weather Prediction.

\begin{table*}[ht]
\centering
\resizebox{0.6\textwidth}{!}{
\begin{tabular}{|l|c|c|}
\hline
    Model & Training Loss & Testing Loss \\ \hline
TFT  & \(0.2630 \) & \(0.9856\) \\ \hline
QTFT (Without QLSTM)  & \(0.2028\) & \(0.8381 \) \\ \hline
QTFT (With QLSTM)  & \(0.1711 \) & \(0.8007\) \\ \hline
\end{tabular}}
\caption{Comparison of training and testing loss values for Axis Bank dataset.}
\label{Table 5}
\end{table*}

\section{Conclusion and Outlook}

We provide a hybrid quantum-classical model architecture for the quantum temporal fusion transformer (QTFT), which is able to perform multi-horizontal time series forecasting.  
In this work, we have explored how large-scale classical learning models are successfully trained and tested on quantum hardware.
We show that under the constraint of a similar architectural structure and an approximately equal number of parameters, the QTFT model slightly performs better than the classical TFT.
Our experimental results demonstrate that the QTFT achieves lower training and testing losses compared to the TFT. Furthermore, incorporating an LSTM component within the QTFT further reduces the overall loss. 
While it is still impractical to run large-scale multi-horizontal time series forecasting due to the limitations of the current quantum simulator software, we emphasize that our architecture is general and scalable in principle.
In general, the quantum variational circuits used in the QTFT model are given broadly and flexibly, incorporating a sufficient number of qubits, more different gate sequences, and a greater number of variational parameters factors that potentially enhance the model's learning capability and higher expressive power.

Finally, if we assume the existence of a perfect quantum device with no noise, deployed with an unlimited number of qubits, exact control mechanisms, and full error-correction, our model has the potential to yield highly efficient and insightful results.

In the future, we are interested in investigating how modifications to the model's architecture could enable its quantum version to achieve better performance than its classical counterpart.    
Specifically, we are interested in closely observing the quantum subcomponents of this model to improve each subcomponent individually, thereby improving the overall performance of the model.


\bibliographystyle{IEEEtran}
\bibliography{references} 

\begin{thebibliography}{10}
\providecommand{\url}[1]{#1}
\csname url@samestyle\endcsname
\providecommand{\newblock}{\relax}
\providecommand{\bibinfo}[2]{#2}
\providecommand{\BIBentrySTDinterwordspacing}{\spaceskip=0pt\relax}
\providecommand{\BIBentryALTinterwordstretchfactor}{4}
\providecommand{\BIBentryALTinterwordspacing}{\spaceskip=\fontdimen2\font plus
\BIBentryALTinterwordstretchfactor\fontdimen3\font minus \fontdimen4\font\relax}
\providecommand{\BIBforeignlanguage}[2]{{%
\expandafter\ifx\csname l@#1\endcsname\relax
\typeout{** WARNING: IEEEtran.bst: No hyphenation pattern has been}%
\typeout{** loaded for the language `#1'. Using the pattern for}%
\typeout{** the default language instead.}%
\else
\language=\csname l@#1\endcsname
\fi
#2}}
\providecommand{\BIBdecl}{\relax}
\BIBdecl

\bibitem{box2015time}
G.~E. Box, G.~M. Jenkins, G.~C. Reinsel, and G.~M. Ljung, \emph{Time series analysis: forecasting and control}.\hskip 1em plus 0.5em minus 0.4em\relax John Wiley \& Sons, 2015.

\bibitem{lim2018forecasting}
B.~Lim, ``Forecasting treatment responses over time using recurrent marginal structural networks,'' \emph{Advances in neural information processing systems}, vol.~31, 2018.

\bibitem{zhang2018multi}
J.~Zhang and K.~Nawata, ``Multi-step prediction for influenza outbreak by an adjusted long short-term memory,'' \emph{Epidemiology \& Infection}, vol. 146, no.~7, pp. 809--816, 2018.

\bibitem{piccialli2021artificial}
F.~Piccialli, F.~Giampaolo, E.~Prezioso, D.~Camacho, and G.~Acampora, ``Artificial intelligence and healthcare: Forecasting of medical bookings through multi-source time-series fusion,'' \emph{Information Fusion}, vol.~74, pp. 1--16, 2021.

\bibitem{kroujiline2016forecasting}
D.~Kroujiline, M.~Gusev, D.~Ushanov, S.~V. Sharov, and B.~Govorkov, ``Forecasting stock market returns over multiple time horizons,'' \emph{Quantitative Finance}, vol.~16, no.~11, pp. 1695--1712, 2016.

\bibitem{capistran2010multi}
C.~Capistr{\'a}n, C.~Constandse, and M.~Ramos-Francia, ``Multi-horizon inflation forecasts using disaggregated data,'' \emph{Economic Modelling}, vol.~27, no.~3, pp. 666--677, 2010.

\bibitem{bose2017probabilistic}
J.-H. B{\"o}se, V.~Flunkert, J.~Gasthaus, T.~Januschowski, D.~Lange, D.~Salinas, S.~Schelter, M.~Seeger, and Y.~Wang, ``Probabilistic demand forecasting at scale,'' \emph{Proceedings of the VLDB Endowment}, vol.~10, no.~12, pp. 1694--1705, 2017.

\bibitem{courty1999timing}
P.~Courty and H.~Li, ``Timing of seasonal sales,'' \emph{The Journal of Business}, vol.~72, no.~4, pp. 545--572, 1999.

\bibitem{salinas2020deepar}
D.~Salinas, V.~Flunkert, J.~Gasthaus, and T.~Januschowski, ``Deepar: Probabilistic forecasting with autoregressive recurrent networks,'' \emph{International journal of forecasting}, vol.~36, no.~3, pp. 1181--1191, 2020.

\bibitem{rangapuram2018deep}
S.~S. Rangapuram, M.~W. Seeger, J.~Gasthaus, L.~Stella, Y.~Wang, and T.~Januschowski, ``Deep state space models for time series forecasting,'' \emph{Advances in neural information processing systems}, vol.~31, 2018.

\bibitem{wen2017multi}
R.~Wen, K.~Torkkola, B.~Narayanaswamy, and D.~Madeka, ``A multi-horizon quantile recurrent forecaster,'' \emph{arXiv preprint arXiv:1711.11053}, 2017.

\bibitem{alaa2019attentive}
A.~M. Alaa and M.~van~der Schaar, ``Attentive state-space modeling of disease progression,'' \emph{Advances in neural information processing systems}, vol.~32, 2019.

\bibitem{makridakis2020m4}
S.~Makridakis, E.~Spiliotis, and V.~Assimakopoulos, ``The m4 competition: 100,000 time series and 61 forecasting methods,'' \emph{International Journal of Forecasting}, vol.~36, no.~1, pp. 54--74, 2020.

\bibitem{li2019enhancing}
S.~Li, X.~Jin, Y.~Xuan, X.~Zhou, W.~Chen, Y.-X. Wang, and X.~Yan, ``Enhancing the locality and breaking the memory bottleneck of transformer on time series forecasting,'' \emph{Advances in neural information processing systems}, vol.~32, 2019.

\bibitem{lim2021temporal}
B.~Lim, S.~{\"O}. Ar{\i}k, N.~Loeff, and T.~Pfister, ``Temporal fusion transformers for interpretable multi-horizon time series forecasting,'' \emph{International Journal of Forecasting}, vol.~37, no.~4, pp. 1748--1764, 2021.

\bibitem{vaswani2017attention}
A.~Vaswani, N.~Shazeer, N.~Parmar, J.~Uszkoreit, L.~Jones, A.~N. Gomez, {\L}.~Kaiser, and I.~Polosukhin, ``Attention is all you need,'' \emph{Advances in neural information processing systems}, vol.~30, 2017.

\bibitem{arute2019quantum}
F.~Arute, K.~Arya, R.~Babbush, D.~Bacon, J.~C. Bardin, R.~Barends, R.~Biswas, S.~Boixo, F.~G. Brandao, D.~A. Buell \emph{et~al.}, ``Quantum supremacy using a programmable superconducting processor,'' \emph{Nature}, vol. 574, no. 7779, pp. 505--510, 2019.

\bibitem{cross2018ibm}
A.~Cross, ``The ibm q experience and qiskit open-source quantum computing software,'' in \emph{APS March meeting abstracts}, vol. 2018, 2018, pp. L58--003.

\bibitem{lanting2014entanglement}
T.~Lanting, A.~J. Przybysz, A.~Y. Smirnov, F.~M. Spedalieri, M.~H. Amin, A.~J. Berkley, R.~Harris, F.~Altomare, S.~Boixo, P.~Bunyk \emph{et~al.}, ``Entanglement in a quantum annealing processor,'' \emph{Physical Review X}, vol.~4, no.~2, p. 021041, 2014.

\bibitem{Harrow_2009}
\BIBentryALTinterwordspacing
A.~W. Harrow, A.~Hassidim, and S.~Lloyd, ``Quantum algorithm for linear systems of equations,'' \emph{Phys. Rev. Lett.}, vol. 103, p. 150502, Oct 2009. [Online]. Available: \url{https://link.aps.org/doi/10.1103/PhysRevLett.103.150502}
\BIBentrySTDinterwordspacing

\bibitem{J__2022}
\BIBentryALTinterwordspacing
A.~J., A.~Adedoyin, J.~Ambrosiano, P.~Anisimov, W.~Casper, G.~Chennupati, C.~Coffrin, H.~Djidjev, D.~Gunter, S.~Karra, N.~Lemons, S.~Lin, A.~Malyzhenkov, D.~Mascarenas, S.~Mniszewski, B.~Nadiga, D.~O’malley, D.~Oyen, S.~Pakin, L.~Prasad, R.~Roberts, P.~Romero, N.~Santhi, N.~Sinitsyn, P.~J. Swart, J.~G. Wendelberger, B.~Yoon, R.~Zamora, W.~Zhu, S.~Eidenbenz, A.~B\"{a}rtschi, P.~J. Coles, M.~Vuffray, and A.~Y. Lokhov, ``Quantum algorithm implementations for beginners,'' \emph{ACM Transactions on Quantum Computing}, vol.~3, no.~4, Jul. 2022. [Online]. Available: \url{https://doi.org/10.1145/3517340}
\BIBentrySTDinterwordspacing

\bibitem{preskill2018quantum}
J.~Preskill, ``Quantum computing in the nisq era and beyond,'' \emph{Quantum}, vol.~2, p.~79, 2018.

\bibitem{gottesman1997stabilizer}
D.~Gottesman, \emph{Stabilizer codes and quantum error correction}.\hskip 1em plus 0.5em minus 0.4em\relax California Institute of Technology, 1997.

\bibitem{gottesman1998theory}
------, ``Theory of fault-tolerant quantum computation,'' \emph{Physical Review A}, vol.~57, no.~1, p. 127, 1998.

\bibitem{biamonte2017quantum}
J.~Biamonte, P.~Wittek, N.~Pancotti, P.~Rebentrost, N.~Wiebe, and S.~Lloyd, ``Quantum machine learning,'' \emph{Nature}, vol. 549, no. 7671, pp. 195--202, 2017.

\bibitem{cerezo2021variational}
M.~Cerezo, A.~Arrasmith, R.~Babbush, S.~C. Benjamin, S.~Endo, K.~Fujii, J.~R. McClean, K.~Mitarai, X.~Yuan, L.~Cincio \emph{et~al.}, ``Variational quantum algorithms,'' \emph{Nature Reviews Physics}, vol.~3, no.~9, pp. 625--644, 2021.

\bibitem{mitarai2018quantum}
K.~Mitarai, M.~Negoro, M.~Kitagawa, and K.~Fujii, ``Quantum circuit learning,'' \emph{Physical Review A}, vol.~98, no.~3, p. 032309, 2018.

\bibitem{wecker2015progress}
D.~Wecker, M.~B. Hastings, and M.~Troyer, ``Progress towards practical quantum variational algorithms,'' \emph{Physical Review A}, vol.~92, no.~4, p. 042303, 2015.

\bibitem{higgott2019variational}
O.~Higgott, D.~Wang, and S.~Brierley, ``Variational quantum computation of excited states,'' \emph{Quantum}, vol.~3, p. 156, 2019.

\bibitem{dunjko2018machine}
V.~Dunjko and H.~J. Briegel, ``Machine learning \& artificial intelligence in the quantum domain: a review of recent progress,'' \emph{Reports on Progress in Physics}, vol.~81, no.~7, p. 074001, 2018.

\bibitem{schuld2015introduction}
M.~Schuld, I.~Sinayskiy, and F.~Petruccione, ``An introduction to quantum machine learning,'' \emph{Contemporary Physics}, vol.~56, no.~2, pp. 172--185, 2015.

\bibitem{huang2021power}
H.-Y. Huang, M.~Broughton, M.~Mohseni, R.~Babbush, S.~Boixo, H.~Neven, and J.~R. McClean, ``Power of data in quantum machine learning,'' \emph{Nature communications}, vol.~12, no.~1, p. 2631, 2021.

\bibitem{clevert2015fast}
D.-A. Clevert, T.~Unterthiner, and S.~Hochreiter, ``Fast and accurate deep network learning by exponential linear units (elus),'' \emph{arXiv preprint arXiv:1511.07289}, 2015.

\bibitem{dauphin2017language}
Y.~N. Dauphin, A.~Fan, M.~Auli, and D.~Grangier, ``Language modeling with gated convolutional networks,'' in \emph{International conference on machine learning}.\hskip 1em plus 0.5em minus 0.4em\relax PMLR, 2017, pp. 933--941.

\bibitem{ba2016layer}
J.~L. Ba, J.~R. Kiros, and G.~E. Hinton, ``Layer normalization,'' \emph{arXiv preprint arXiv:1607.06450}, 2016.

\bibitem{gal2016theoretically}
Y.~Gal and Z.~Ghahramani, ``A theoretically grounded application of dropout in recurrent neural networks,'' \emph{Advances in neural information processing systems}, vol.~29, 2016.

\bibitem{wang2018high}
M.~Wang, S.~Lu, D.~Zhu, J.~Lin, and Z.~Wang, ``A high-speed and low-complexity architecture for softmax function in deep learning,'' in \emph{2018 IEEE asia pacific conference on circuits and systems (APCCAS)}.\hskip 1em plus 0.5em minus 0.4em\relax IEEE, 2018, pp. 223--226.

\bibitem{hochreiter1997long}
S.~Hochreiter and J.~Schmidhuber, ``Long short-term memory,'' \emph{Neural computation}, vol.~9, no.~8, pp. 1735--1780, 1997.

\bibitem{schuld2021effect}
M.~Schuld, R.~Sweke, and J.~J. Meyer, ``Effect of data encoding on the expressive power of variational quantum-machine-learning models,'' \emph{Physical Review A}, vol. 103, no.~3, p. 032430, 2021.

\bibitem{lloyd2020quantumembeddingsmachinelearning}
\BIBentryALTinterwordspacing
S.~Lloyd, M.~Schuld, A.~Ijaz, J.~Izaac, and N.~Killoran, ``Quantum embeddings for machine learning,'' 2020. [Online]. Available: \url{https://arxiv.org/abs/2001.03622}
\BIBentrySTDinterwordspacing

\bibitem{bergholm2018pennylane}
V.~Bergholm, J.~Izaac, M.~Schuld, C.~Gogolin, S.~Ahmed, V.~Ajith, M.~S. Alam, G.~Alonso-Linaje, B.~AkashNarayanan, A.~Asadi \emph{et~al.}, ``Pennylane: Automatic differentiation of hybrid quantum-classical computations,'' \emph{arXiv preprint arXiv:1811.04968}, 2018.

\bibitem{fingerhuth2018open}
M.~Fingerhuth, T.~Babej, and P.~Wittek, ``Open source software in quantum computing,'' \emph{PloS one}, vol.~13, no.~12, p. e0208561, 2018.

\bibitem{benedetti2019parameterized}
M.~Benedetti, E.~Lloyd, S.~Sack, and M.~Fiorentini, ``Parameterized quantum circuits as machine learning models,'' \emph{Quantum science and technology}, vol.~4, no.~4, p. 043001, 2019.

\bibitem{schuld2020circuit}
M.~Schuld, A.~Bocharov, K.~M. Svore, and N.~Wiebe, ``Circuit-centric quantum classifiers,'' \emph{Physical Review A}, vol. 101, no.~3, p. 032308, 2020.

\bibitem{kandala2017hardware}
A.~Kandala, A.~Mezzacapo, K.~Temme, M.~Takita, M.~Brink, J.~M. Chow, and J.~M. Gambetta, ``Hardware-efficient variational quantum eigensolver for small molecules and quantum magnets,'' \emph{nature}, vol. 549, no. 7671, pp. 242--246, 2017.

\bibitem{farhi2014quantum}
E.~Farhi, J.~Goldstone, and S.~Gutmann, ``A quantum approximate optimization algorithm,'' \emph{arXiv preprint arXiv:1411.4028}, 2014.

\bibitem{mcclean2016theory}
J.~R. McClean, J.~Romero, R.~Babbush, and A.~Aspuru-Guzik, ``The theory of variational hybrid quantum-classical algorithms,'' \emph{New Journal of Physics}, vol.~18, no.~2, p. 023023, 2016.

\bibitem{havlivcek2019supervised}
V.~Havl{\'\i}{\v{c}}ek, A.~D. C{\'o}rcoles, K.~Temme, A.~W. Harrow, A.~Kandala, J.~M. Chow, and J.~M. Gambetta, ``Supervised learning with quantum-enhanced feature spaces,'' \emph{Nature}, vol. 567, no. 7747, pp. 209--212, 2019.

\bibitem{farhi2018classification}
E.~Farhi and H.~Neven, ``Classification with quantum neural networks on near term processors,'' \emph{arXiv preprint arXiv:1802.06002}, 2018.

\bibitem{dallaire2018quantum}
P.-L. Dallaire-Demers and N.~Killoran, ``Quantum generative adversarial networks,'' \emph{Physical Review A}, vol.~98, no.~1, p. 012324, 2018.

\bibitem{chen2020variational}
S.~Y.-C. Chen, C.-H.~H. Yang, J.~Qi, P.-Y. Chen, X.~Ma, and H.-S. Goan, ``Variational quantum circuits for deep reinforcement learning,'' \emph{IEEE access}, vol.~8, pp. 141\,007--141\,024, 2020.

\bibitem{mari2020transfer}
A.~Mari, T.~R. Bromley, J.~Izaac, M.~Schuld, and N.~Killoran, ``Transfer learning in hybrid classical-quantum neural networks,'' \emph{Quantum}, vol.~4, p. 340, 2020.

\bibitem{chen2022quantum}
S.~Y.-C. Chen, S.~Yoo, and Y.-L.~L. Fang, ``Quantum long short-term memory,'' in \emph{Icassp 2022-2022 IEEE international conference on acoustics, speech and signal processing (ICASSP)}.\hskip 1em plus 0.5em minus 0.4em\relax IEEE, 2022, pp. 8622--8626.

\bibitem{schuld2019quantum}
M.~Schuld and N.~Killoran, ``Quantum machine learning in feature hilbert spaces,'' \emph{Physical review letters}, vol. 122, no.~4, p. 040504, 2019.

\bibitem{qiskit2024}
A.~Javadi-Abhari, M.~Treinish, K.~Krsulich, C.~J. Wood, J.~Lishman, J.~Gacon, S.~Martiel, P.~D. Nation, L.~S. Bishop, A.~W. Cross, B.~R. Johnson, and J.~M. Gambetta, ``Quantum computing with {Q}iskit,'' 2024.

\bibitem{du2022quantum}
Y.~Du, T.~Huang, S.~You, M.-H. Hsieh, and D.~Tao, ``Quantum circuit architecture search for variational quantum algorithms,'' \emph{npj Quantum Information}, vol.~8, no.~1, p.~62, 2022.

\bibitem{ayoub2025high}
F.~Ayoub and J.~D. Baeder, ``High-entanglement capabilities for variational quantum algorithms: the poisson equation case: F. ayoub, jd baeder,'' \emph{Quantum Information Processing}, vol.~24, no.~8, p. 229, 2025.

\bibitem{li2024quantum}
G.~Li, X.~Zhao, and X.~Wang, ``Quantum self-attention neural networks for text classification,'' \emph{Science China Information Sciences}, vol.~67, no.~4, p. 142501, 2024.

\bibitem{schuld2019evaluating}
M.~Schuld, V.~Bergholm, C.~Gogolin, J.~Izaac, and N.~Killoran, ``Evaluating analytic gradients on quantum hardware,'' \emph{Physical Review A}, vol.~99, no.~3, p. 032331, 2019.

\bibitem{paszke2019pytorch}
A.~Paszke, ``Pytorch: An imperative style, high-performance deep learning library,'' \emph{arXiv preprint arXiv:1912.01703}, 2019.

\end{thebibliography}
\end{document}